# Real-Time Multiple Object Tracking

## A Study on the Importance of Speed

SAMUEL MURRAY

# Abstract

Multiple object tracking consists of detecting and identifying objects in video. In some applications, such as robotics and surveillance, it is desired that the tracking is performed in real-time. This poses a challenge in that it requires the algorithm to run as fast as the frame-rate of the video. Today's top performing tracking methods run at only a few frames per second, and can thus not be used in real-time. Further, when determining the speed of the tracker, it is common to not include the time it takes to detect objects. We argue that this way of measuring speed is not relevant for robotics or embedded systems, where the detecting of objects is done on the same machine as the tracking. We propose that one way of running a method in real-time is to not look at every frame, but skip frames to make the video have the same frame-rate as the tracking method. However, we believe that this will lead to decreased performance.

In this project, we implement a multiple object tracker, following the *tracking-by-detection* paradigm, as an extension of an existing method. It works by modelling the movement of objects by solving the filtering problem, and associating detections with predicted new locations in new frames using the Hungarian algorithm. Three different similarity measures are used, which use the location and shape of the bounding boxes. Compared to other trackers on the MOTChallenge leaderboard, our method, referred to as *C++SORT*, is the fastest non-anonymous submission, while also achieving decent score on other metrics. By running our model on the Okutama-Action dataset, sampled at different frame-rates, we show that the performance is greatly reduced when running the model – including detecting objects – in real-time. In most metrics, the score is reduced by $50\%$, but in certain cases as much as $90\%$. We argue that this indicates that other, slower methods could not be used for tracking in real-time, but that more research is required specifically on this.

## Sammanfattning

För att spåra rörliga objekt i video (eng: *multiple object tracking*) krävs att man lokaliserar och identifierar dem. I vissa tillämpningar, såsom robotik och övervakning, kan det krävas att detta görs i realtid, vilket kan vara svårt i praktiken, då det förutsätter att algoritmen kan köras lika fort som videons bildfrekvensen. De kraftfullaste algoritmerna idag kan bara analysera ett fåtal bildrutor per sekund, och lämpar sig därför inte för realtidsanvändning. Dessutom brukar tiden per bildruta inte inkludera den tid det tar att lokalisera objekt, när hastigheten av en algoritm presenteras. Vi anser att det sättet att beräkna hastigheten inte är lämpligt inom robotik eller inbyggda system, där lokaliseringen och identifiering av objekt sker på samma maskin. Många algoritmer kan köras i realtid genom att hoppa över det antal bildrutor i videon som krävs för att bildfrekvensen ska bli densamma som algoritmens frekvens. Dock tror vi att detta leder till sämre prestanda.

I det här projektet implementerar vi en algoritm för att identifiera rörliga objekt. Vår algoritm bygger på befintliga metoder inom paradigmen *tracking-by-detection* (ung. spårning genom detektion). Algoritmen uppskattar hastigheten hos varje objekt genom att lösa ett filtreringsproblem. Utifrån hastigheten beräknas en förväntad ny position, som kopplas till nya observationer med hjälp av Kuhn–Munkres algoritm. Tre olika likhetsmått används, som på olika sätt kombinerar positionen för och formen på objekten. Vår metod, *C++SORT*, är den snabbaste icke-anonyma metoden publicerad på MOTChallenge. Samtidigt presterar den bra enligt flera andra mått. Genom att testa vår algoritm på video från Okutama-Action, med varierande bildfrekvens, kan vi visa att prestandan sjunker kraftigt när hela modellen – inklusive att lokalisera objekt – körs i realtid. Prestandan enligt de flesta måtten sjunker med 50%, men i vissa fall med så mycket som 90%. Detta tyder på att andra, långsammare metoder inte kan användas i realtid, utan att mer forskning, specifikt inriktad på spårning i realtid, behövs.

# Acknowledgments

This thesis was done while I was an Invited Researcher at the National Institute of Informatics in Tokyo, Japan. As generous financial aid for my stay, I was awarded the Gadelius scholarship of the Sweden-Japan Foundation, for which I am very grateful. I would also like to thank the people at Prendinger laboratory, for making the stay an enjoyable one.

# Contents

# Chapter 1

# Introduction

The fields of image and video analysis have gained a lot of attention over the last years, in part due to the success of deep learning models. Most state-of-the-art methods use convolutional neural networks (CNNs) – deep networks which have shown to produce good results without the need for manual feature extraction. The shift of attention towards CNNs for image analysis happened in 2012, when Krizhevsky et al. [1] convincingly won the ILSVRC (ImageNet Large-Scale Visual Recognition Challenge) [2], a competition used for benchmarking image analysis models. The challenge was in *image classification*, where the goal is to label each image with one of multiple classes – *e.g.* to tell if a picture is of a cat or a dog. More accurate models have since been created, and in 2015 computers could outperform humans in this task [3]. The use of CNNs has not been limited to image recognition, but has been successfully adopted in tasks such as *object detection* (localising objects of given classes), *semantic segmentation* (labelling each pixel in the image), and *image captioning* (producing a descriptive text of the content) [4, 5].

Video analysis is closely related to the field of image analysis, a video being multiple images stacked in time. Similar challenges as in image analysis have been tackled, including *video classification* and *object detection in video*, but also tasks exclusive to video, such as *object tracking* (identifying objects across multiple frames), *trajectory prediction* (estimating paths of objects) and *action recognition* (classifying actions in a video sequence). As in image analysis, CNNs have produced great results for these tasks [6, 7].

Having accurate image and video analysis models is fundamental to autonomous vehicles and robots. Object detection models can enable self-driving cars to avoid collisions, and some level of scene understanding is required for robots to react to their surroundings. With the increased use of unmanned aerial vehicles (UAVs) – commonly known as *drones* – for tasks such as surveillance, delivery, and search and rescue [8], it is important to develop tools suitable for these applications and adapted to UAVs. Often, it is required to identify and track humans and other moving objects. For example, in surveillance tasks, it can be essential to track humans in order to detect unusual behaviour. Unfortunately, not all video analysis models are usable on UAVs; the most powerful models require high-end computational hardware, but for UAVs that should be light and inexpensive, the hardware is necessarily restricted.

## 1.1   Problem definition

When a video contains multiple moving objects that we wish to track, we refer to this as *multiple object tracking*. See Figure 1.1 for an illustration. In some sense, the task is an extension of object detection, since in addition to detecting objects, we need to connect detections between frames to get a consistent tracking. Object detection is still an unsolved problem, and the most powerful methods are limited by their speed [9]. Adding tracking capabilities on top of the detector usually slows down the algorithm further. Because of this, multiple object tracking is difficult to do in real-time, since the best algorithms can only analyse a few frames per second at best, even on powerful hardware [10]. For such algorithms to run in real-time, it would be necessary to skip multiple frames in order to prevent an ever-increasing delay. To the best of our knowledge, not much work has been done to investigate how sampling of frames would affect the tracking performance. Indeed, in MOTChallenge, a yearly competition and benchmark of multiple object tracking, there is little incentive to make models capable of running in real-time.

In this project, we wish to investigate how real-time multiple object tracking would work. To do this, a multiple object tracking system is implemented, which is evaluated on existing datasets to enable comparison with other work. Since the goal is for the method to run in real-time, an important factor is the speed of the tracker. Because of this, it is decided that a simple approach is preferred over a slow but potentially more powerful one. By comparing with work on public leaderboards, we will get an objective evaluation of our method. To get an understanding on how more powerful models would perform if run in real-time by sampling frames, our model is run with different configurations and different sampling rates on a large dataset. The effect on performance from decreasing the frame-rate is measured, with the hope of gaining insight on how other models would fare under such circumstances.

As a use case for real-time tracking, one can imagine that the complete model is deployed on small UAVs with restricted hardware. The UAVs communicate information wirelessly to a ground control station, and due to the low bandwidth and to prevent latency, video footage is not transferred. Instead, all computations are done on the UAVs, and the extracted information is sent as messages at regular intervals. The system is required to work in real-time and in an online fashion, *i.e.* to give an output steadily and without much delay. This is particularly challenging given the restricted hardware of the UAVs, which prevents the usage of memory and computing heavy models.

In MOTChallenge, the users are provided with text files of detections for each frame in the dataset, as a way to ensure that everyone is on equal ground, and that the tracking capability is actually measured, as opposed to the detection quality. This motivates the use of a paradigm called *tracking-by-detection* (see Section 2.2), which is agnostic to the type of detector used. Our method is also tested on Okutama-Action [11], which has the benefit of providing two pre-trained models, one for object detection and one for action recognition, along with the detections they output as text files. By modifying the output of the action recognition model slightly, it can be used for normal object detection. The benefit of having two different models is that the tracking performance can be measured when provided detections of different quality.

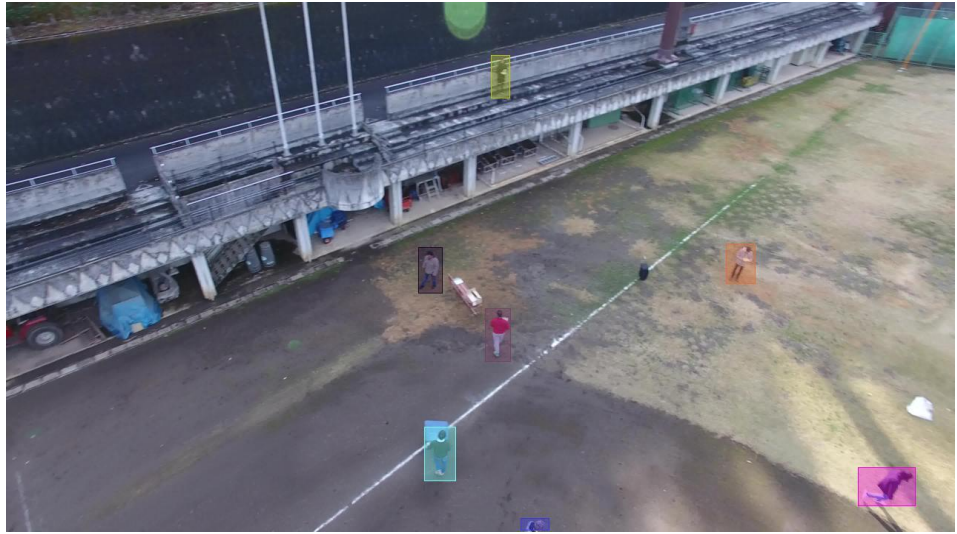

(a) Initial frame

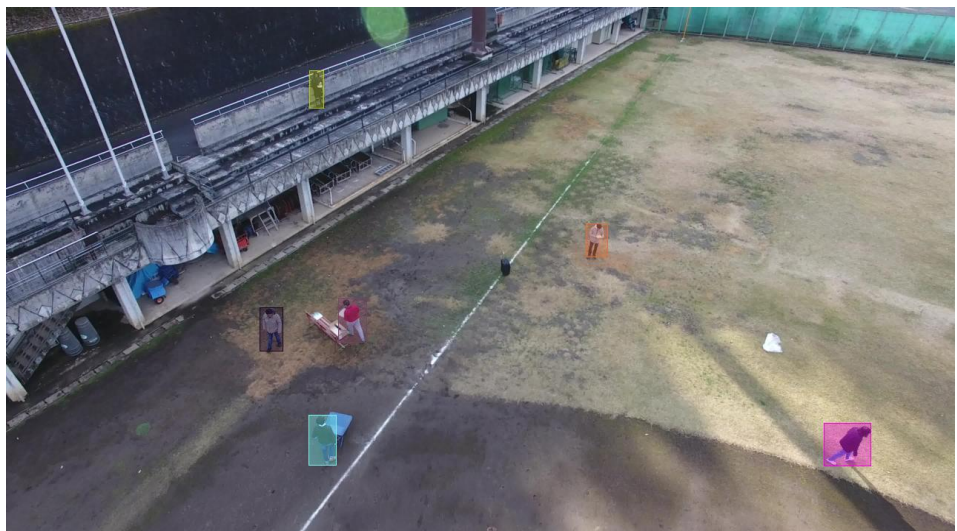

(b) 50 frames (1.7 s) later

Figure 1.1: Tracking of seven persons in a video (one of which leaves the scene between the displayed frames). Each person is enclosed in a coloured bounding box, with the colours representing identities. Potential failures include mixing up the identities (colours), or missing some persons completely.

## 1.2 Research question

In this thesis, we will investigate if existing online, tracking-by-detection models can successfully be used to track multiple objects in real-time. To answer this, we will implement a tracking system and evaluate its performance under different configurations. To further investigate real-time tracking, we will also measure how the frame-rate of the video affects the ability to track, since skipping frames is a way to make most models work in real-time. The hope is that the findings can guide future research on real-time tracking, or at least increase the interest in this problem. The thesis will be considered successful either if we find a model that can be used for real-time multiple object track-
ing, or if we gain insights into what is required in order to achieve this.

## 1.3   Limitations

Since the goal is to have real-time tracking, only tracking methods with low inference
speed are considered. This means that methods using appearance features to associate
objects between frames are not considered, since computing such features is relatively
slow. Instead, only geometrical and positional properties are used. Additionally, to achieve
the modularity required to use either object detection or action recognition models, no
model that requires a specific type of object detector is considered. For the same reason,
methods that incorporate the predicted location from the tracking to guide or improve
the object detector are not used. Last, the aim of the thesis is not to propose a new track-
ing model, and thus we will use and combine ideas from existing methods and attempt
to implement them in an efficient way.

## 1.4   Outline

The thesis is organised as follows: Chapter 2 introduces the reader to relevant theory,
and presents previous work on multiple object tracking. The exact details of the approach
of this project, which builds on the theory of the preceding chapter, along with the exper-
iments run, are presented in Chapter 3. The results of the experiments are given in Chap-
ter 4, and are compared to that of previous work. Chapter 5 concludes the thesis with a
discussion on the performance of the model, and ideas for future work.

# Chapter 2

# Related work

This chapter introduces the reader to the task of tracking, the challenges that exist and previously proposed methods to solve them. After a general description of the topic in Section 2.1, the idea of *tracking-by-detection* is presented in Section 2.2, which lays the foundation for our work. A brief discussion on real-time tracking is held in Section 2.3. Last, commonly used datasets as well as evaluation metrics are listed in Section 2.4 and Section 2.5 respectively.

## 2.1  Multiple object tracking

As the name suggests, multiple object tracking consists of keeping track of objects in a video as they move around. A formal description would be: *for each frame in a video, localise and identify all objects of interest, so that the identities are consistent throughout the video*. In other words, a good model has to accurately detect objects in each frame, and provide a consistent labelling of them. See Figure 1.1 for an illustration. Challenges arise when objects are partially or completely occluded, or temporarily leave the field of view, since ideally the objects should keep their former IDs when reappearing. Furthermore, objects whose paths intersect might confuse the model and cause it to erroneously switch their IDs.

Different scenarios exist that allow for different types of models. An important distinction is that of *online* versus *offline* models [12]. An online model receives video input on a frame-by-frame basis, and has to give an output for each frame. This means that, in addition to the current frame, only information from past frames can be used. Offline models, on the other hand, have access to the entire video, which means that information from both past and future frames can be used. The task can then be viewed as an optimisation problem, where the goal is to find a set of paths that minimise some global loss function. This has been solved using linear programming [13, 14] and k-shortest path optimisation methods [15]. Since offline trackers have access to more information, one can expect better performance from these models. It should be noted, however, that real-time usage requires online models, as future frames are obviously unavailable. For a discussion on real-time models, see Section 2.3. Unless otherwise stated, the rest of this chapter is dedicated to online models.

Models can be further categorised into *single class* and *multiple class* models [16]. In the former, all tracked objects are of the same class, *e.g. person*, while in the latter, multiple classes are present, *e.g. pedestrian*, *bicycle* and *car*. On one hand, having multiple classes introduces the problem of classifying each object. On the other hand, this could act as a distinguishing feature when the paths of objects of separate classes intersect.

An extensive survey from 2006 of object tracking systems [17] showed that research until then had focused on finding good features, object representations and motion models to improve the tracking quality. It was noted that most methods assumed some restriction on the setting that greatly facilitated the tracking. For example, by assuming a static camera, it was possible to find all moving objects by applying background subtraction [18]. Other common assumptions were that objects had a high contrast with respect to the background, moved smoothly and with little to no occlusion. Yilmaz et al. [17] identified these assumptions to be greatly limiting when it comes to real-world application, stating that while said assumptions increased the performance, they made the models too specific.

## 2.2    Tracking-by-detection

The survey of Yilmaz et al. [17] was published before deep learning gained attention in image analysis in 2012, when Krizhevsky et al. [1] won the ILSVRC using deep learning. Today's convolutional neural networks (CNNs) greatly outperform previous methods in the task of image classification. An overview of CNNs is presented in Section 2.2.1. For object detection, three datasets commonly used for benchmarking are ImageNet [2], Pascal VOC [19] and MS COCO [20]. The best performing models for all three challenges use CNNs [21, 22, 23], indicating that the detection capability today has far surpassed that of 2006. This in turn has lead to a renewed interest in a particular tracking paradigm, called *tracking-by-detection*.

Although all tracking systems need to detect the objects at some stage [17], tracking-by-detection methods make a clear distinction between the detection and tracking of objects. The general idea is to for each frame, first localise all objects using an object detector, then associate detected objects between frames using features such as location and appearance. Avidan [24] was early to use this approach, by using a pre-trained support vector machine (SVM) to detect vehicles, and optical flow-like equations to connect detections between frames. It is noted by Bewley et al. [25] and Yu et al. [26], however, that the performance of tracking-by-detection models is heavily dependent on the accuracy of the detection model, which indicates that CNN-based object detectors could greatly boost the tracking performance, even with simple tracking models. As discussed in Section 2.3 however, having a powerful detector can come at the cost of reduced frame-rate in real-time scenarios.

A common methodology is to split the tracking into two phases: prediction of object location, and matching of detections and predictions [25]. That is, for each new frame, the complete tracking model does the following: (1) *Detect* objects of interest, (2) *Predict* new locations of objects from previous frames, (3) *Associate* objects between frames by similarity of detected and predicted locations. Although some methods diverge from this approach slightly, *e.g.* by matching objects in multiple steps based on confidence values [27], or keeping multiple hypotheses active [28, 29], most works differ simply doing these three steps in distinct ways.

### 2.2.1 Detection

The goal in object detection is, as the name suggests, to detect all objects of a certain class in an image. Alternatively, multiple classes can exist, in which case it is also required to correctly classify each object. The input to an object detector is an image, and the output is a list of bounding boxes, with labels in the case of multiple classes. A bounding box is usually represented by the pixel coordinates of the top-left and bottom-right corner of the bounding box, or the coordinates of the centre as well as the width and height of the box. Furthermore, most object detectors provide a confidence value with each box, indicating how trustworthy the detection is. A common performance measure of an object detector is the mean average precision over all classes (see Section 2.5). As stated above, object detection methods using CNNs are state-of-the-art, outperforming previous methods such as SVMs.

**Convolutional neural network** A CNN is a type of artificial neural network, designed to take advantage of structures in images and similar data. In 1990, LeCun et al. [30] used a small CNN to recognise handwritten digits – this was the first successful use of CNNs. The network, named LeNet, consisted of 5 layers: 2 sets of a *convolutional* layer followed by a *pooling* layer, and finally a *fully connected* layer. The architecture of LeNet is shown in Figure 2.1. Though modern CNNs are much deeper (*i.e.* have more layers), the general structure of alternating convolutional and pooling layers, followed by fully connected layers, remains [4].

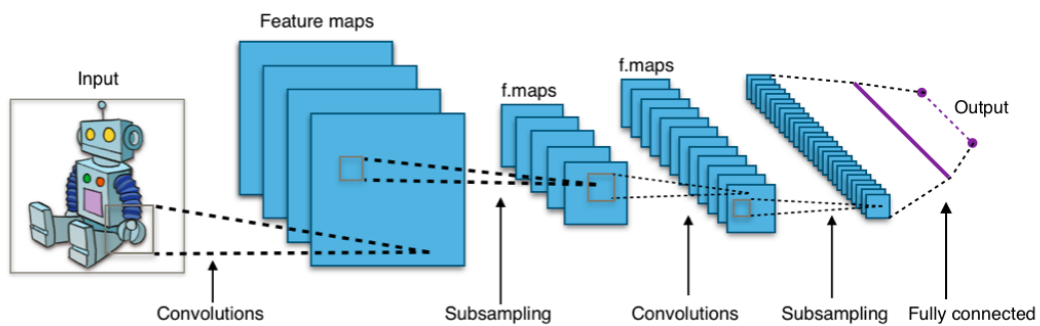

Figure 2.1: A convolutional neural network, with 5 layers: 2 convolutional layers, each followed by a pooling layer, and finally a fully connected layer.[1]

Fully connected layers have a weight associated with each pair of input and output. The output is computed as the inner product of the weights and the input. These weights have to be learned, and since the number of weights is proportional to the number of inputs and to the number of outputs, the number of parameters quickly grows for fully connected layers. Convolutional layers, on the other hand, have a small set of weights referred to as *filters*. To compute the output, each filter is swept across the input, and at each position the inner product between the filter and the input at that location is computed. This has the benefit of sharing weights, which drastically reduces the number of parameters. Additionally, once a filter has learned to capture a certain structure, it is able to find it anywhere in the image, since the filter will be applied at all positions. This is desirable for images, since the same objects and structures could appear anywhere. Last, pooling layers are used to reduce the dimensionality of the data. Like convolutional

---

[1] https://commons.wikimedia.org/wiki/File:Typical_cnn.png, Accessed: May 20, 2017

layers, pooling is done at each position of the input, though usually with no overlap. Different variants exist, but in max pooling, which is the most commonly used, the output is computed as the maximum value of the input. Notably, pooling layers have no learnable parameters [4].

Popular well-performing CNN architectures used for object detection are R-FCN [31], Faster R-CNN [32] and SSD [33].

### 2.2.2 Prediction

A multitude of approaches have been suggested for how to predict new locations of tracked objects. Some examples include computing the optical flow[2] to determine the new position [34], or using recurrent neural networks [35, 36], Kalman filters [25, 37, 16] or particle filters [38, 27, 26] to model the velocity of objects, and with this predict the position in future frames.

**Filtering problem** Modelling the velocity by only looking at the position at discrete time steps is an example of the filtering problem. At each time step $t$, the goal is to model the state $x_t$ given potentially noisy observations $z_{1:t}$. A visualisation of the process is given in Equation (2.1). In the case of tracking-by-detection, the observations are the detected bounding boxes in each frame. To model the movement of objects, the hidden state contains information about the position and the velocity, which has to be inferred by looking only at the observations.

$$
\begin{array}{ccccccccc}
x_0 & \to & x_1 & \to & \cdots & \to & x_{k-1} & \to & x_k \\
\downarrow & & \downarrow & & & & \downarrow & & \downarrow \\
z_0 & & z_1 & & & & z_{k-1} & & z_k
\end{array}
\tag{2.1}
$$

To predict the position in a upcoming frame $k + 1$ equates to predicting the value of $z_{k+1}$. Knowing the time difference between $k$ and $k + 1$, an estimated new position is obtained by adding the velocity of $x_k$ (which is estimated by solving the filtering problem) to $z_k$ (which is given by the object detector). More information can be added to $x$, such as acceleration, and change in size and shape of the bounding box, though this is done at the cost of increased complexity. Two methods to solve the filtering problem are presented below.

**Kalman filter** The Kalman filter is a way of optimally estimating the state of a linear dynamical system. Kalman [39] presented the theory in 1960 and showed that the estimation minimises the mean of the squared error, given that the noise is normally distributed.

Assume we have a state $x \in \mathbb{R}^n$ and a control signal $u \in \mathbb{R}^l$. Let the state be governed by the equation

$$
x_k = Ax_{k-1} + Bu_{k-1} + w_{k-1},
\tag{2.2}
$$

where $A$ is the state transition model, $B$ is the control-input model and $w_k$ is the process noise. Further, assume the state is not directly measurable, but is related to the observable variable $z \in \mathbb{R}^m$ as

$$
z_k = Hx_k + v_k,
\tag{2.3}
$$

---

[2]This is computationally expensive, which makes it unsuited for real-time applications.

where $H$ is the observation model and $v_k$ is the observation noise. The random variables $w_k$ and $v_k$ are assumed to be independent and normally distributed with zero mean and covariance $Q$ and $R$ respectively. The Kalman filter computes the optimal state estimate $\hat{x}$ by recursively combining previous estimates with new observations. It consists of two distinct phases: *predict*, where the optimal state $\hat{x}_k^-$ prior to observing $z_k$ is computed; and *update*, where after observing $z_k$, the optimal posterior state $\hat{x}_k$ is computed. Additionally, it computes the prior and posterior estimate error covariance $P_k^{(\cdot)} = E[e_k^{(\cdot)}e_k^{(\cdot)T}]$, where $e_k^{(\cdot)} = x_k^{(\cdot)} - \hat{x}_k^{(\cdot)}$. The prediction step is

$$\hat{x}_k^- = A\hat{x}_{k-1} + Bu_{k-1}, \tag{2.4}$$

$$P_k^- = AP_{k-1}A^T + Q, \tag{2.5}$$

and the update step is

$$K_k = P_k^- H^T (HP_k^- H^T + R)^{-1}, \tag{2.6}$$

$$\hat{x}_k = \hat{x}_k^- + K_k(z_k - H\hat{x}_k^-), \tag{2.7}$$

$$P_k = (I - K_k H)P_k^-, \tag{2.8}$$

where $I$ is the identity matrix, $Q$ is the process noise covariance and $R$ is the measurement noise covariance. $K_k$ is called the Kalman gain, and influences how much the observation impacts the state estimate. In addition to specifying the process matrices $A$, $B$, $Q$, $H$ and $R$, initial estimates of $\hat{x}$ and $P$ have to be provided when using a Kalman filter [40].

**Particle filter** Assume we have a hidden state $x \in \mathbb{R}^n$ and an observation $z \in \mathbb{R}^m$ that are governed by the following equations:

$$x_k = f(x_{k-1}) + w_{k-1} \tag{2.9}$$

$$z_k = h(x_k) + v_k. \tag{2.10}$$

Here, $f$ and $h$ are (potentially nonlinear) known functions, and $w$ and $v$ are noise, mutually independent and with known probability density functions. Particle filters can be used to approximate the probability distribution $p(x_t|z_{1:t})$ online in a nonlinear dynamical system, such as the one above [41]. It can be noted that if $f$ and $h$ are linear, and the noise is Gaussian, then Equations (2.9) and (2.10) can be solved optimally using a Kalman filter (compare with Equations (2.2) and (2.3)).

To approximate $p(x_t|z_{1:t})$, a set of $N$ *particles* is used, each of which represents a sample from the distribution. Increasing the number of particles leads to more accurate approximations, but also more computations. The algorithm consists of two steps, *predict* and *update*, similar to the Kalman filter. In the predict step, each particle is updated using Equation (2.9), giving $N$ samples $\{x_k^i\}_{i=1}^N$ corresponding to $p(x_t|z_{1:t-1})$. Subsequently in the update step, each particle is given a weight $w_k^i = p(z_k|x_k^i)$ indicating how well it corresponds to the observation $z_k$. After normalising the weights, a new set of $N$ particles is obtained by sampling with replacement from $\{x_t^i\}_{i=1}^N$, using the weights to determine the probability of selecting a particle. The new set of particles approximates $p(x_t|z_{1:t})$. The initial positions of the particles need to be specified, distributed *e.g.* normally around the first observation, or uniformly in the space of possible states [42].

### 2.2.3 Association

The association task consists of determining what detection corresponds to what object, based on the predictions from the previous step, or alternatively if a detection represents a new object. Problems arise when tracked objects leave the scene or are otherwise not detected, new objects enter the scene, the detector produces false positives – *i.e.* "detecting" an object that does not exist – or when the predicted positions differ greatly from reality.

**Assignment problem** In the case where the number of detections is equal to the number of tracked objects, this is a case of the assignment problem, in which the goal is to find an optimal matching between two sets of object. Commonly, one set is called *agents* and the other *tasks*. There is a cost $c_{ij}$ associated with assigning a task $j$ to an agent $i$, and the objective is to find an assignment with minimum total cost, such that no agent is assigned more than one task, and no task is assigned to more than one agent. Formally, if $A$ is the set of agents, $T$ the set of tasks, and $x_{ij}$ represents the assignment, taking value $1$ if task $j$ is assigned to agent $i$ and $0$ otherwise, then the objective is to minimise

$$\sum_{i \in A} \sum_{j \in T} c_{ij} x_{ij}, \tag{2.11}$$

subject to

$$\sum_{i \in A} x_{ij} = 1, j \in T, \tag{2.12}$$

$$\sum_{j \in T} x_{ij} = 1, i \in A, \tag{2.13}$$

$$x_{ij} \geq 0, i \in A, j \in T. \tag{2.14}$$

One can also consider the same problem, but with the goal of maximising the cost rather than minimising it.

**Hungarian algorithm** The Hungarian algorithm [43], also known as Kuhn–Munkres algorithm, solves the assignment problem in polynomial time, with time complexity $\mathcal{O}(n^3)$ where $n$ is the number of agents (and tasks) [44]. The input to the algorithm is a cost matrix $C$, where $C(i, j)$ is the cost $c_{ij}$, and in four steps the matrix is manipulated to give the optimal matching. In the case where the number of agents is not equal to the number of tasks, $C$ can be padded with rows or columns with large values to make it square. Inversely, if the goal is to find a matching with maximum cost, the matrix can be padded with zeros. Then in the final matching, assignments corresponding to added rows or columns are discarded.

**Similarity measures** In order to compute the cost matrix $C$ between detections and predictions, we need to define what it means for two bounding boxes to be (dis)similar. A simple way is to compute the intersection over union (IoU), also known as the Jaccard index, between two bounding boxes $A$ and $B$ as

$$\text{IoU}(A, B) = \frac{|A \cap B|}{|A \cup B|} = \frac{|A \cap B|}{|A| + |B| - |A \cap B|}, \tag{2.15}$$

where $|\cdot|$ denotes the area. This has the benefit of combining distance and similarity in size and shape in a nice way, but unfortunately it gives $0$ for any two bounding boxes that do not intersect. This means that if the predictions are too inaccurate, all values in the cost matrix could be 0, meaning that all assignments are equally good. Despite this, IoU has successfully been used either as the only measure [25] or in tandem with appearance information [37].

More sophisticated tracking models combine different similarity measures to calculate the cost, most of which can loosely be separated into three categories: *distance* measures, *shape* measures, and *appearance* measures. These are summed or multiplied with different weights, yielding a single value indicating the overall similarity between bounding boxes. Distance and shape measures look only at the position and geometry of the bounding boxes, whereas appearance features are calculated from the actual pixels within each bounding box.

Below, let $A$ and $B$ be two bounding boxes and $C(A, B)$ be the affinity between them. Further, let $W$, $H$ and $(X, Y)$ be the width, the height and the centre coordinate of a bounding box respectively. Sanchez-Matilla et al. [27] propose the following affinity measure, where a low value indicates similarity:

$$C(A, B) = c_{dist}(A, B) * c_{shp}(A, B),  \tag{2.16}$$

where

$$c_{dist}(A, B) = \frac{\sqrt{(X_A - X_B)^2 + (Y_A - Y_B)^2}}{Q_{dist}},  \tag{2.17}$$

$$c_{shp}(A, B) = \frac{\sqrt{(H_A - H_B)^2 + (W_A - W_B)^2}}{Q_{shp}}.  \tag{2.18}$$

The denominators $Q_{dist}$ and $Q_{shp}$ have the values of the diagonal and the area of the image respectively, and work to normalise the two factors. The two parts are multiplied instead of summed, to promote bounding boxes that are similar both in shape and in position, rather than in only one aspect. The factors are the Euclidean distance of the position and the shape respectively, which one could argue makes the formula intuitive and easy to understand.

Measuring the difference in position and shape exponentially, and also using appearance features, Yu et al. [26] define the cost as:

$$C(A, B) = c_{app}(A, B) * c_{dist}(A, B) * c_{shp}(A, B),  \tag{2.19}$$

where

$$c_{app}(A, B) = \cos(feat_A, feat_B),  \tag{2.20}$$

$$c_{dist}(A, B) = e^{-w_1 * ((\frac{X_A - X_B}{W_A})^2 + (\frac{Y_A - Y_B}{H_A})^2)},  \tag{2.21}$$

$$c_{shp}(A, B) = e^{-w_2 * (\frac{|H_A - H_B|}{H_A + H_B} + \frac{|W_A - W_B|}{W_A + W_B})}.  \tag{2.22}$$

In the non-symmetrical formula for $c_{dist}$, $A$ is assumed to be a detection, and $B$ a prediction. The vectors *feat* in $c_{app}$ are extracted by a CNN with a 128-dimensional output. The weights $w_1$ and $w_2$ are set to 0.5 and 1.5 respectively. Compared to the two other similarity measures, this formula is more complicated, and thus the output might not be easy to interpret. However, it has the benefit of always giving an output between 0 and 1, indicating no and complete similarity respectively.

Simpler appearance features have also been proposed, such as colour histograms and HOG features [45]. Though using appearance features can lead to increased performance, it has the downside of longer computation times.

## 2.3 Real-time tracking

There is no clear definition in multiple object tracking on what inference speed is required for a model to be said to work in real-time. Instead, it is said that a model is real-time if it can give output as fast as, or faster than, the rate of which input is given [46]. That is, a tracker is real-time if it can analyse 60 s of video in less than one minute – otherwise, it is not. However, the video is not continuous, but is captured at a certain frame-rate, and the frames are input to the tracker with a constant interval. This means that a 60 s video captured at 30 FPS results in 1800 ($60 \times 30$) images to be processed by the tracker, but decreasing the frame-rate to 10 FPS gives only 600 ($60 \times 10$) images. Since online methods are required to give an output before the next input is provided, the delay of the system (*i.e.* the time it takes to get an output after providing an input) is directly related to the speed in which images are analysed. For example, if a tracker can analyse images in 30 FPS, then the delay is $1/30$ s. Of course, additional delay can arise from the time it takes to send the output to the user, though this overhead can be expected to be constant, and is not a part of the tracker.

Arguably, for some real-world applications of tracking, it is not required to analyse 30 frames each second; we can adjust the frame-rate of the input to the speed of the tracker and say that anything above 1 FPS (*i.e.* a delay of less than 1 s) is acceptable. Of course, in certain applications a shorter delay might be necessary, and thus a higher frame-rate would be required. One way to achieve real-time performance is to skip all incoming frames until the tracker is done analysing a given frame, so that the input is always the most recently captured one. This way, the delay is constant, and we can say that the tracker runs in real-time. However, the lower the frame-rate, the farther we can expect objects to have moved between frames. This indicates that decreasing the frame-rate makes tracking harder, since it is easier to predict where an object will be 0.1 s from now, rather than 1 s. As seen in the leaderboard of MOTChallenge [10], most current research on online tracking requires a low frame-rate on the input in order to run in real-time. In fact, only one submission runs faster than 30 FPS, but its accuracy is less than $2/3$ of that of the best models, see Table 4.3. However, for the other models, we expect the performance to decrease when the input is sampled to make them run in real-time, as argued above.

Since in tracking-by-detection it is assumed that detections are provided independently of the tracker, most research only presents the speed of the tracker. With the definition given above, the trackers can be said to be real-time if they can process more frames each second than the frame-rate of the video [14, 25, 37], which is correct under the assumption that the detections can be provided at this speed. However, in machines that need to both produce the detections and compute the tracking, as is the case for embedded systems, the resources are necessarily shared between the two tasks, which will decrease the frame-rate at which the system can operate in real-time. Thus, when determining what frame-rate a tracker can run at, the time it takes for the detector to analyse an image should be added to that of the tracker. This means that there is a trade-off when designing the detector; a more powerful detector gives more accurate detections, which facilitates the tracking, but at the same time runs slower [9, 47] which makes tracking harder since more frames must be skipped. Unfortunately, very little work has been done on investigating this trade-off, with the models submitted to MOTChallenge always considering every frame, independent of the speed of the trackers.

## 2.4  Datasets

Multiple benchmarks exist for evaluating tracking models, the most common of which are presented below.

**MOTChallenge** [48] MOTChallenge is a yearly competition used to benchmark multiple object tracking models. The dataset – video sequences labelled with bounding boxes for each pedestrian – is collected from multiple sources, which differ in resolution, frame rate, illumination *etc*. There are two variations of the challenge, each with a public leaderboard online to promote competition and comparison of models. In one task, you are only given the raw video sequences, and thus need to both detect and track the objects; in the other you are given the sequences along with a set of detections, and the task is to make an as accurate tracker as possible using these detections. Since the provided detections are not state-of-the-art, the results of the first task is generally better. However, the second task is arguably more interesting, since it measures only the trackers, when having access to the same detections. Submissions are compared using multiple metrics (see Section 2.5), and a tool is provided to compute these metrics for any data. There is a large number of competitors each year, which makes this challenge ideal for evaluating your work. A sample frame is shown in Figure 2.2.

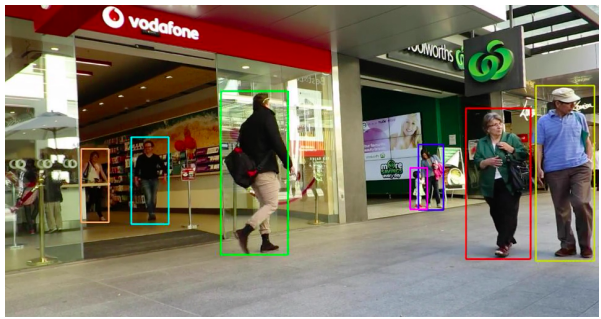

Figure 2.2: Sample frame from MOTChallenge. Ground truth bounding boxes are displayed. Colours represent IDs.

**ImageNet VID** [2] Having previously been a competition in image recognition, ImageNet now features challenges in object detection in images and video, as well as object tracking. The sequences contain labelled objects from 30 different categories. Mean average precision is used as the only metric, which makes comparison with other methods more limited. An example frame with instances of the class *bicycle* is shown in Figure 2.3.

**Okutama-Action** [11] A recently published dataset is Okutama-Action, which consists of aerial-view video captured from UAVs. It was designed to feature challenges likely to exist in real-world applications of UAVs, including abrupt camera movement, different view-angles and varying altitude. Because of the altitude, which ranges from 10 to 45 meters, the area of interest is generally small, see Figure 2.4 for a sample frame. Okutama-Action features only pedestrians, but provides two distinct annotations: one for object tracking and one for action recognition. The videos are of high resolution (3840x2160) and contain up to 10 concurrent humans. For the task of action recognition, 12 action classes are available, a few of which are displayed in Figure 2.5. Since the dataset is recently published, no previous results are available for comparison.

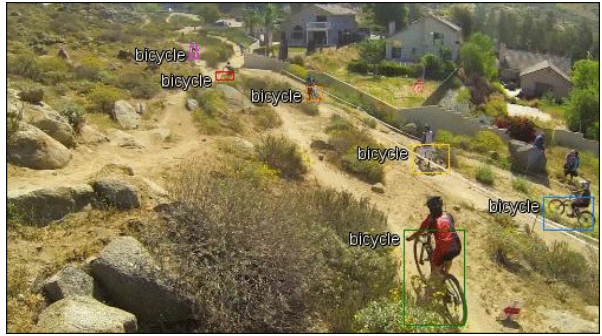

Figure 2.3: Sample frame from ImageNet. Ground truth bounding boxes are displayed, along with the class labels. Colours represent IDs.

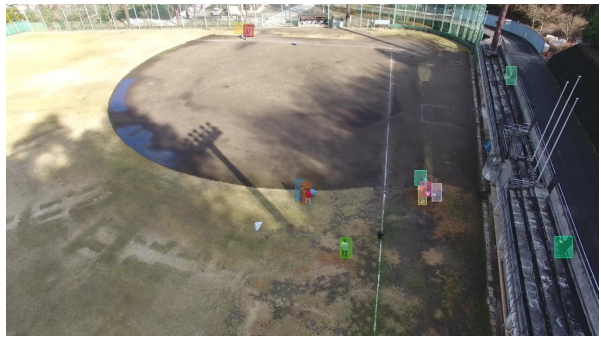

Figure 2.4: Sample frame from Okutama-Action with human detection annotation. Ground truth bounding boxes are displayed. Colours represent IDs.

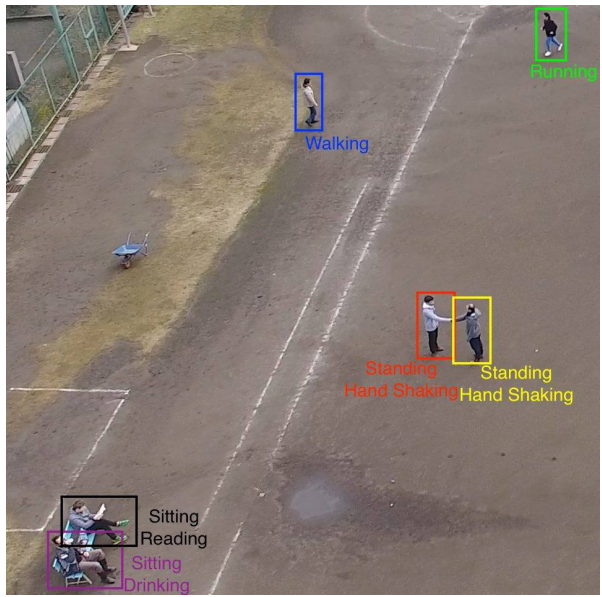

Figure 2.5: Cropped sample frame from Okutama-Action with action detection annotation. Ground truth bounding boxes are displayed, along with the actions performed. Colours represent IDs.

## 2.5   Evaluation metrics

Common ways to measure the quality of object detection models include calculating the *accuracy* and the *mean average precision* (mAP) at given IoU thresholds. That is, to calculate the accuracy at IoU 0.5, you count detections as correct if they have IoU $\geq 0.5$ with the ground truth, and incorrect otherwise. As stated earlier, these values can heavily influence the performance of tracking-by-detection models, and thus comparing two trackers that use different detectors is not as meaningful. However, even with the same detection accuracy and precision, the quality can be vastly different between different tracking models, so additional metrics are needed.

As made obvious in the previous section, different datasets use different metrics to compare multiple object tracking models. However, in 2008 Bernardin and Stiefelhagen [49] proposed two standardised metrics to be used for multiple object tracking, referred to as the *CLEAR MOT metrics*. These metrics, *MOTA* and *MOTP*, along with the additional metrics used in MOTChallenge, are presented below.

**FP** False positives is the total number of occurrences where an object is detected although no object exists.

**FN** False negatives is the total number of occurrences where an existing object is not detected.

**ID Sw.** Identity switches is the number of times an object is assigned a new ID in its track.

**MOTA** Multiple object tracking accuracy is a combination of three errors: it is computed as

$$\text{MOTA} = 1 - \frac{\sum_t (fn_t + fp_t + id\_sw_t)}{\sum_t g_t}, \tag{2.23}$$

where for frame $t$, $g_t$ is the number of objects present, $fn_t$ the number of false negatives, $fp_t$ the number of false positives, and $id\_sw_t$ the number of identity switches. A perfect tracker has MOTA $= 100\%$.

**MOTP** Multiple object tracking precision measures the alignment of the predicted bounding boxes and the ground truth. Notably, it does not take into account any tracking performance such as keeping consistent trajectories, but is heavily reliant on the detection quality. The original formula was

$$\text{MOTP} = \frac{\sum_{i,t} d_t^i}{\sum_t c_t}, \tag{2.24}$$

where $d_t^i$ is the distance between the ground truth and prediction of object $i$, and $c_t$ the number of matches found in frame $t$. However, this is slightly modified in MOTChallenge so that a perfect tracker has MOTP $= 100\%$.

**FAF** False alarm per frame is the average number of false positives in each frame.

**MT** Mostly tracked targets is the number of ground-truth tracks that are assigned the same label for at least $80\%$ of the video.

**ML** Mostly lost targets is the number of ground-truth tracks that are assigned the same label for at most 20% of the video.

**Frag** Fragmentation counts the number of times an object is lost in a frame but then re-detected in a future frame, thus fragmenting the track.

**Hz** The speed of a tracker is measured in Hz (frames per second). Unless otherwise stated, only the tracking is timed, *i.e.* the speed of the detector is not included.

# Chapter 3

# Method

The work of this project is split into two distinct but related parts. The first is to implement a competitive multiple object tracker, with emphasis on speed. Our model, which we refer to as *C++SORT*, is described in Section 3.1. To evaluate the tracker, the datasets from MOTChallenge (see Section 2.4) are used, mainly because of the public leaderboards which facilitate comparison to other work. The exact configuration of the tracker used on these datasets is presented in Section 3.2.

Second, we investigate how the frame-rate of the video influences the performance of the tracker. This is interesting, since most state-of-the-art methods are not able to run in real-time (see Section 2.3). We answer this by running our method with different configurations on the same video sequences, sampled at multiple different frame-rates. Additionally, we do a case study where we run our full model, including object detection, in real-time. A more detailed description of this is given in Section 3.3.

## 3.1   C++SORT

Our work follows the *tracking-by-detection* paradigm, which implies that the task of detecting objects in each frame is done separately from the tracking. Usually, as in MOTChallenge, we are not concerned with obtaining the detections, but assume that we have some way of extracting them, and use the detections as input to the tracking algorithm. To speed up the construction and deployment of tracking-by-detection models, the detections are often computed offline before-hand, and are stored as text files for fast feeding to the tracker. Note that in our work, the tracker makes no assumption on the nature of the detections even in the case where they are obtained online, nor does it attempt to guide the detector based on current trackings.

We use the method of Bewley et al. [25], called *SORT – Simple Online and Realtime Tracking*. We do, however, extend the method slightly by considering other prediction methods than Kalman filters and other similarity measures than the IoU. To distinguish the methods and their results, we refer to our model as *C++SORT*, which also reveals the choice of language (see Section 3.4). Much of the relevant theory is presented in Section 2.2 and its subsections.

In short, the idea is to keep track of each object by modelling its movement across the frame with a *predictor*, which could be *e.g.* a Kalman filter or a particle filter (see Section 2.2.2). For each frame, the following happens: objects are detected, new locations of already tracked objects are predicted using the predictors, and the detected and tracked objects are matched based on the similarities of the bounding boxes. The predictors are then updated with their associated detections, and the estimated positions are returned as trackings. Additionally, new predictors are initiated for any unmatched detection, and unused predictors are removed. The method is presented in more rigour in Algorithm 1.

---

**Algorithm 1** C++SORT

---

Set values for *threshold*, *maxAge* and *minHitStreak*
**for all** *frames* **do**
    Get *detections* from *frame*                                           ▷ Section 3.1.1
    Remove detections with confidence less than *threshold*
    Predict new locations of objects using *predictors*                ▷ Section 3.1.2
    Associate *predictors* and *detections* based on similarity      ▷ Section 3.1.3
    **for all** *predictors* **do**
        **if** *predictor* has associated *detection* **then**
            PROPAGATE(*predictor*, *detection*)
        **else**
            PROPAGATE(*predictor*)
        **end if**
    **end for**
    **for all** *unmatchedDetections* **do**
        Initialise new *predictor* with *detection*
    **end for**
    Remove predictors that have had no associated detection the last *maxAge* frames
    **for all** *predictors* **do**
        **if** *predictor* had associated detection this and the last *minHitStreak* frames **then**
            Add state of *predictor* to *trackings*
        **end if**
    **end for**
    **return** *trackings*
**end for**

---

### 3.1.1 Detection

In MOTChallenge, detections are provided as text files, with each row containing the following information on one detection: frame number, x-position, y-position, width, height and confidence (ranging from 0 to 1). Note that the provided detections are not perfect, meaning that the positions can be slightly off, and that some false positives and false negatives are expected. The confidence value is used to determine which detections to use in the tracker. By a small grid search, 0.4 was found to be a good value to use as threshold in Algorithm 1. In addition to the information stated above, the true ID is provided in the ground truth, which is obviously only available for the training set.

For Okutama-Action, a type of CNN called SSD [33] is trained and used to detect objects. In the task of human detection, it has only one output, but for action recognition it has one for each action type. However, since humans are the only actors, and every human is labelled with an action each frame, the output of the action recognition network can be used for tracking, if the action label is ignored. The network trained for action recognition is slower and less accurate than the other one, since action recognition is a more difficult task than human detection. The networks along with text files with the detections are publicly available. Additionally, we have access to the ground truth of both the training and the test set. For more details on the networks, including how they were trained, see the work of Barekatain et al. [11].

### 3.1.2  Prediction

Two ways of predicting new positions are investigated, namely Kalman filters and particle filters, the theory of which are presented in Section 2.2.2. Each predictor tracks a single object, which it does by modelling its movement with a filter. In addition to this, it keeps track of some meta data such as how long the object has been visible, or for how long the predictor has been inactive. Unfortunately, the use of particle filters is not successful, and due to a much slower run speed, no extensive experiments are run with this type of predictor.

For either approach, the state $x$ contains information on the centre pixel coordinate, the area, the width-height ratio, the velocity and the change in area over time, $\frac{d(\text{area})}{dt}$, of a bounding box. The observation $z$ on the other hand only has the values which can be observed in a single frame, namely the centre coordinate, the area and the width-height ratio of each bounding box. The area and ratio pair is preferred over storing the width and height, since the width-height ratio is assumed to be fairly constant throughout a video sequence. This is also the reason why the change in ratio over time is not modelled in $x$.

Getting the predicted new location of an object in frame $k + 1$ equates to computing $x_{k+1}$ (and then the observation variable $z_{k+1}$) in the equations for the dynamic systems, Equations (2.2), (2.3), (2.9) and (2.10). However, all variables in $z$ are contained in $x$, so calculating $z$ is done by simply removing the fields exclusive to $x$. Since we do not have a control signal $u$, the computation for the Kalman filter is $x_{k+1} = Ax_k$. For the particle filter, it is a bit more complicated. The state is represented by a set of $N$ particles, each of which has its own guess of the true values. To get the predicted new location, each particle has to be propagated one step, similarly to the Kalman filter. After that, however, the particles need to be combined into a single measure. For simplicity, we use an unweighted average, but one could also consider using the particles' weights when computing the mean. The weight we use when resampling the particles is the IoU of the particles state and the bounding box of the detected object. This way, particles that are close to detected objects get a high weight, whereas particles that do not intersect with the bounding box are removed.

After the association step, which is described in the next section, the predictors should propagate their states to the next time step. If a predictor has an associated detection, it is updated as described in Section 2.2.2. Otherwise, $x_{k+1}$ is computed without adjusting for any observation, *i.e.* only the prediction step for the Kalman filter. When using a particle filter, noise is added to the propagation of each particle, since there is an increased uncertainty regarding the tracked object's true position.

As a very primitive baseline, a third model, referred to as *stationary predictor*, is used when running the experiments on Okutama-Action. As the name implies, this model assumes that all objects are stationary, *i.e.* it does not model their velocities. The intuition is that this predictor should suffer more from decreased frame-rates than the Kalman fil-ter and the particle filter, since objects will have moved a greater distance in between frames.

### 3.1.3   Association

As described in Section 2.2.3, the goal in the association step is to find an optimal match between predictors and detections. The quality of a match is determined by the sum of the cost of each associated predictor-detection pair. Depending on if a high cost indicates similarity or dissimilarity, the summed cost should be maximised or minimised respectively.

The first step when finding an optimal match is generating the cost matrix. This is a matrix where each element $c_{i,j}$ is the (dis)similarity between predictor $i$ and detection $j$. Given the cost matrix, the Hungarian algorithm can find the optimal match between predictors and detections. It is, however, beneficial to have a threshold value used to reject associations if the cost is too high or too low. Otherwise, if two consecutive frames contain two persons – but in the first frame only one person is detected, and in the second frame only the other is detected – the method would believe the two detections to be the same person, when it should be obvious that this is not the case given the distance between the two.

Three similarity measures are compared, referred to as *IoU cost*, *linear cost* and *exponential cost*. Notably, all measures only use the geometrical and positional properties of the bounding boxes, *i.e.* no information of the appearance or colour histograms are used. This decision was made since such measures would slow down the computation, or introduce unwanted dependencies on the object detector, if information from its hidden layers were to be used.

The IoU cost is precisely the IoU between the predicted and the detected bounding boxes, as given in Equation (2.15):

$$C_{IoU}(A, B) = \text{IoU}(A, B). \tag{3.1}$$

For this measure, a high value indicates similarity, and a threshold of $0.3$ was found to give good results. This measure does have the downside that bounding boxes with no intersection get the similarity value $0$, independent of how far away they are, and thus it can not be known how dissimilar they are more than that they do not intersect. This is however overruled by the fact that a threshold is used, so all bad matches are removed anyway.

Using the formula proposed by Sanchez-Matilla et al. [27], the linear cost is similar to Equation (2.16). However, for implementation reasons the cost is inverted to become

$$C_{lin}(A, B) = \frac{Q_{dist}}{\sqrt{(X_A - X_B)^2 + (Y_A - Y_B)^2}} \cdot \frac{Q_{shp}}{\sqrt{(H_A - H_B)^2 + (W_A - W_B)^2}}. \tag{3.2}$$

With this modification, this measure shares the property of the other two that a high value indicates similarity. It was found that $10000$ is a good value to use as threshold.

The third measure, the exponential cost, is the one proposed by Yu et al. [26] in Equation (2.19), but without the use of appearance features. We thus get

$$C_{exp}(A, B) = e^{-w_1 * ((\frac{X_A - X_B}{W_A})^2 + (\frac{Y_A - Y_B}{H_A})^2)} \cdot e^{-w_2 * (\frac{|H_A - H_B|}{H_A + H_B} + \frac{|W_A - W_B|}{W_A + W_B})}. \tag{3.3}$$

Again, a value close to 1 indicates similarity, and 0.5 was used as threshold to discard dissimilar bounding boxes.

## 3.2   MOTChallenge

For the MOTChallenge, there are two tracking datasets available, called *2D MOT 2015* (from now on referred to as *MOT15*) and *MOT16*. By evaluating on the training set, it was found that the IoU cost was the best for this dataset, though the exponential cost gave similar results. A small grid search was used to find optimal values on parameters in Algorithm 1, such as thresholds, and maximum age and minimum hit streak of predictors. The thresholds are the ones presented previously in this chapter, *maxAge* was set to 1 and *minHeatStreak* to 3. After deciding on models and parameter values, the method was run on the test set, and the output was submitted to MOTChallenge. The results, along with the results of other groups, are presented in Section 4.1. Notably, this is the same configuration as used by Bewley et al. [25]. However, they have only evaluated the model with custom detections, making it harder to compare with other work, since the quality of detections can heavily influence the tracking performance. Furthermore, we expect a faster run time due to our choice of programming language (see Section 3.4).

## 3.3   Tracking with different frame-rates

The second part of this project, as stated in the beginning of this chapter, is to investigate how the performance of the tracker is affected by the frame-rate of the input video. Since the video sequences in MOTChallenge are diverse in terms of resolution and frame-rate, it was decided to not run these experiments on that dataset, but instead run them on the Okutama-Action dataset. Because all sequences in Okutama-Action have the same frame-rate, and are more similar to each other in general, it is expected that there will not be a few sequences which are more heavily affected by change in frame-rate, thereby skewing the average performance. Furthermore, Okutama-Action has the advantage that we have access to two different object detectors and their detections, as well as the ground truth for both the training and the test set.

The parameters are set to the same values as for MOTChallenge. This was decided on after running a small grid search on the training set. Each experiment session consists of running a particular configuration on the test set with seven different frame-rates: 30, 15, 10, 6, 3, 2 and 1. Since the videos are captured in 30 FPS, a frame-rate of 30 means that all frames are used, while a frame-rate of 1 means that only every 30th frame is used. In total, 18 different configurations are used (three different detection sets, three different similarity measures, and two different predictors). The detections are from the human detection model, the action recognition and the ground truth for the task of object detection. Notably, for the ground truth, no information about the IDs are used, *i.e.* it corresponds to having a perfect object detector. The similarity measures are the ones mentioned previously – the IoU, the linear and the exponential cost functions. Since no good results are obtained with particle filters, only the Kalman filter and the stationary predictor are used. The main results are given in Section 4.2, while complete tables of all sessions are found in Appendix A.

### 3.3.1   Detecting and tracking in real-time

As discussed in Section 2.3, the frame-rate in which a tracker can run in real-time is affected not only by the speed of the tracker, but also by the speed of the detector. As a case study, the full model – that is, detecting objects online with the use of CNNs rather than loading the detection text files – is run on the Okutama-Action dataset. Rather than using a fixed frame-rate, the number of frames skipped depends on the speed of the model, to make sure that there is no cumulative delay. In a real life scenario, the model would be fed the last recorded frame as soon as it is done computing the current one; in our case, we measure the time it takes for the model to analyse each frame, and skip the corresponding number of frames. Unfortunately, when feeding frames in this manner we are not able to use the MOTChallenge DevKit to measure the performance, so instead the average frame-rate is measured and the results of the other experiments are used to see what values this corresponds to. To simulate the use case of deploying the full model on an embedded system, this experiment is run on a Jetson TX1 Developer Kit, which, thanks to its size, could be mounted on a UAV to analyse the video as it is captured. The outcome of this experiment is presented in Section 4.2.1.

## 3.4   Environment

Our work is done in C++, instead of in Python as the work of Bewley et al. [25]. This is necessary, since the environment in which our code is to be integrated is written in C++, and thus using another language would lead to undesired overhead. Furthermore, C++ is expected to result in faster run time compared to languages such as Python.

The deep learning models are done in Caffe [50], which has native support for C++, and GPU computations using CUDA. In an attempt to make the code easier to deploy on new machines, only two external libraries are used in addition to Caffe: dlib [51] and Boost.

Some scripts used for visualisation and pre-processing of data are made in Python. The code of this project, along with instructions on how to deploy it, is available at `https://github.com/samuelmurray/tracking-by-detection`.

MOTChallenge provides a *DevKit* with Matlab scripts used to evaluate a tracker. The output of the tracker is saved as text files, and these are input to the Matlab scripts. The input is compared to the ground truth, which also has to be provided, and the different metrics are calculated and presented.

The experiments are, unless otherwise stated, run on a 2012 MacBook Air, with a 1.8 GHz Intel Core i5, and Intel HD Graphics 4000.

# Chapter 4

# Results

In this chapter, the results of the different experiments are presented. The results on MOT-Challenge are presented in Section 4.1, and those from the suite of experiments run with different configurations, at different frame-rates, are shown in Section 4.2. Due to the large number of metrics used and the number of experiments run, only the most important results are presented in this chapter, while complete tables are found in Appendix A.

## 4.1  MOTChallenge

The model described in Chapter 3 was run on the datasets of the previous two years of MOTChallenge, MOT15 and MOT16. Different configurations of the model were run on the training set, to see which were most effective. The results of the top performing configurations are shown in Tables 4.1 and 4.2 for the two years respectively. For both datasets, using a Kalman filter with the exponential or the IoU cost proved most successful. Additionally, for MOT15, we ran the same configurations with the ground truth as input, instead of the provided detections. As can be seen in the table, this lead to a major increase in performance, giving near perfect tracking.

|  | MOTA | MOTP | FAF | MT | ML | FP | FN | ID Sw. | Frag |
|---|---|---|---|---|---|---|---|---|---|
| *Detections* |  |  |  |  |  |  |  |  |  |
| IoU cost | 25.8 | 72.5 | 1.24 | 62 | 235 | 6797 | 22013 | 782 | 1176 |
| Exponential cost | 23.1 | 72.4 | 1.28 | 48 | 232 | 7034 | 22632 | 1006 | 1382 |
| *Ground truth* |  |  |  |  |  |  |  |  |  |
| IoU cost | 95.8 | 94.8 | 0.00 | 348 | 45 | 4 | 1643 | 33 | 38 |
| Exponential cost | 96.0 | 94.7 | 0.00 | 361 | 35 | 2 | 1535 | 40 | 38 |

Table 4.1: Results of using using Kalman filters with two different similarity measures on the MOT15 training set. The upper half is from using the provided detections; the bottom half is from using the ground truth as input.

Though by a small margin, the IoU cost outperformed the exponential cost (at least when using the normal detections) and was thus used for the test set of MOT15 and MOT16. The results are presented in Tables 4.3 and 4.4. Also in these tables are results of other groups, taken from the online leaderboards [10]. Only groups that use the provided detections and whose methods are online (as opposed to offline, see Section 2.1) are shown, since comparison to other methods is not meaningful in our case. Additionally, only results of groups who have made their code available or described their work in a report (*i.e.* no *anonymous* submissions to MOTChallenge) are displayed. This is done, since there is no way of confirming or analysing the results of anonymous groups. The methods are ordered by their average rank across all metrics, which is provided on the leaderboards. This is done in order not to emphasise any one metric.

|  | MOTA | MOTP | FAF | MT | ML | FP | FN | ID Sw. | Frag |
|---|---|---|---|---|---|---|---|---|---|
| *Detections* | | | | | | | | | |
| IoU cost | 27.4 | 78.8 | 0.59 | 22 | 302 | 3130 | 76202 | 786 | 1165 |
| Exponential cost | 27.3 | 78.8 | 0.61 | 22 | 303 | 3234 | 76137 | 837 | 1240 |

Table 4.2: Results of using using Kalman filters with two different similarity measures on the MOT16 training set. The provided detections are used as input.

|  | MOTA | MOTP | FAF | MT | ML | FP | FN | ID Sw. | Frag | Hz |
|---|---|---|---|---|---|---|---|---|---|---|
| TDAM [52] | 33.0 | **72.8** | 1.7 | 13.3 | 39.1 | 10064 | 30617 | 464 | 1506 | 5.9 |
| MDPNN [53] | **37.6** | 71.7 | 1.4 | **15.8** | **26.8** | 7933 | **29397** | 1026 | 2024 | 1.9 |
| SCEA [54] | 29.1 | 71.1 | 1.0 | 8.9 | 47.3 | 6060 | 36912 | 604 | 1182 | 6.8 |
| CDA_DDALpb [55] | 32.8 | 70.7 | **0.9** | 9.7 | 42.2 | **4983** | 35690 | 614 | 1583 | 2.3 |
| MDP [34] | 30.3 | 71.3 | 1.7 | 13.0 | 38.4 | 9717 | 32422 | 680 | 1500 | 1.1 |
| oICF [56] | 27.1 | 70.0 | 1.3 | 6.4 | 48.7 | 7594 | 36757 | 454 | 1660 | 1.4 |
| LDCT [57] | 4.7 | 71.7 | 2.4 | 11.4 | 32.5 | 14066 | 32156 | 12348 | 2918 | 20.7 |
| OMT_DFH [58] | 21.2 | 69.9 | 2.3 | 7.1 | 46.5 | 13218 | 34657 | 563 | 1255 | 28.6 |
| GMPHD_15 [59] | 18.5 | 70.9 | 1.4 | 3.9 | 55.3 | 7864 | 41766 | **459** | 1266 | 19.8 |
| EAMTTpub [27] | 22.3 | 70.8 | 1.4 | 5.4 | 52.7 | 7924 | 38982 | 833 | 1485 | 12.2 |
| GSCR [60] | 15.8 | 69.4 | 1.3 | 1.8 | 61.0 | 7597 | 43633 | 514 | **1010** | 28.1 |
| C++SORT | 21.7 | 71.2 | 1.5 | 3.7 | 49.1 | 8422 | 38454 | 1231 | 2005 | **1112.1** |
| TSDA_OAL [61] | 18.6 | 69.7 | 2.8 | 9.4 | 42.3 | 16350 | 32853 | 806 | 1544 | 19.7 |
| RNN_LSTM [35] | 19.0 | 71.0 | 2.0 | 5.5 | 45.6 | 11578 | 36706 | 1490 | 2081 | 165.2 |
| RMOT [62] | 18.6 | 69.6 | 2.2 | 5.3 | 53.3 | 12473 | 36835 | 684 | 1282 | 7.9 |
| TC_ODAL [63] | 15.1 | 70.5 | 2.2 | 3.2 | 55.8 | 12970 | 38538 | 637 | 1716 | 1.7 |

Table 4.3: Leaderboard of the MOT15 test set, from MOTChallenge. The best value in each category is boldface. The method of this report, C++SORT, is highlighted in grey.

|  | MOTA | MOTP | FAF | MT | ML | FP | FN | ID Sw. | Frag | Hz |
|---|---|---|---|---|---|---|---|---|---|---|
| MDPNN16 [53] | **47.2** | 75.8 | **0.5** | **14.0** | **41.6** | **2681** | **92856** | 774 | 1675 | 1.0 |
| oICF [56] | 43.2 | 74.3 | 1.1 | 11.3 | 48.5 | 6651 | 96515 | **381** | 1404 | 0.4 |
| GMPHP_HDA [59] | 30.5 | 75.4 | 0.9 | 4.6 | 59.7 | 5169 | 120970 | 539 | **731** | 13.6 |
| CDA_DDALv2 [55] | 43.9 | 74.7 | 1.1 | 10.7 | 44.4 | 6450 | 95175 | 676 | 1795 | 0.5 |
| C++SORT | 31.5 | **77.3** | **0.5** | 4.3 | 59.9 | 3048 | 120278 | 1587 | 2239 | **687.1** |
| EAMTT_pub [27] | 38.8 | 75.1 | 1.4 | 7.9 | 49.1 | 8114 | 102452 | 965 | 1657 | 11.8 |
| OVBT [64] | 38.4 | 75.4 | 1.9 | 7.5 | 47.3 | 11517 | 99463 | 1321 | 2140 | 0.3 |

Table 4.4: Leaderboard of the MOT16 test set, from MOTChallenge. The best value in each category is boldface. The method of this report, C++SORT, is highlighted in grey.

Our work is called *C++SORT*, and is highlighted in grey. The best score in each metric is highlighted in bold. In both of the challenges, our work vastly outperformed the competitors in speed (Hz). Also, in MOT16 it was the best method in terms of MOTP and FAF. However, in the same year it was ranked last when measuring the number of mostly tracked (MT) and mostly lost (ML) paths, as well as the number of ID switches and fragmentations. In MOT15, however, our work arguably fell in the middle ground, not being the worst in any metric, but also only winning in terms of speed. As a final note, *SORT* [25], for which results using the provided detections are not available, runs at $260.5$ FPS on MOT15, indicating that the choice of programming language can heavily affect the speed.

## 4.2  Tracking with different frame-rates

Our model was run on the test set of Okutama-Action using 18 different configurations, each run while sampling the video at seven different intervals. A sampling interval of $i$ means that every $i$th frame is used, and corresponds to a frame-rate of $1/i$ FPS. The performance in terms of MOTA, MOTP and percentage of paths not labelled "mostly lost" for the different frame-rates are shown in Figures 4.1 to 4.3. Each plot contains nine configurations: the three similarity measures – Exponential, IoU and Linear – and the three different detection sets – from the human detection model (Human), from the action recognition model (Action) and the ground truth (GT). The similarity measures are distinguished by colour and marker style, while the detections have different line styles. Numerical values of all metrics provided by the MOTChallenge DevKit are found in Appendix A.

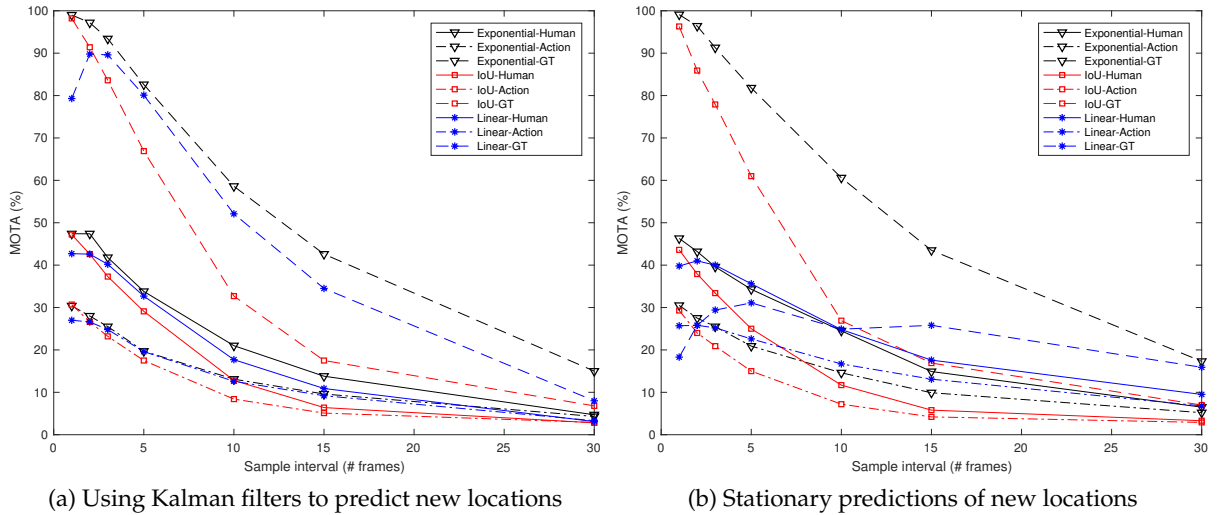

(a) Using Kalman filters to predict new locations        (b) Stationary predictions of new locations

Figure 4.1: MOTA for different similarity measures – the exponential cost in black triangle, IoU cost in red square, linear cost in blue asterisk – and different detection types – human detection in solid line, action recognition in dash-dot line, ground truth in dashed line – shown as functions over sample interval.

In general, for all metrics, the tracker performed better with more accurate detections. It is also noted that the MOTA and the percentage of tracked paths drops rapidly with a decreased frame-rate. In general, using Kalman filters boosted the performance slightly compared to using stationary predictions, though not consistently. Figure 4.2 shows that

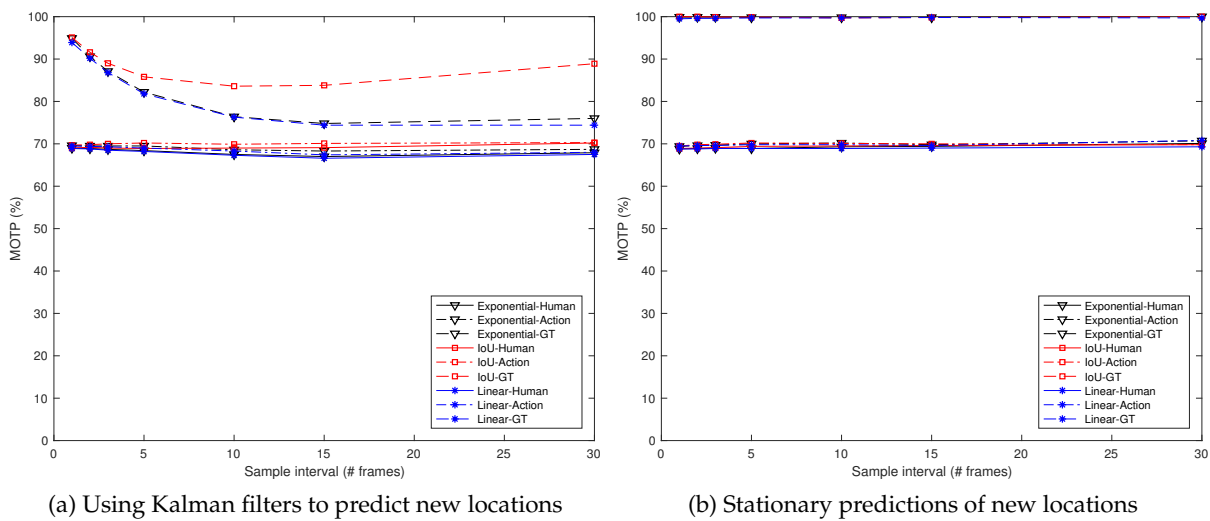

(a) Using Kalman filters to predict new locations     (b) Stationary predictions of new locations

Figure 4.2: MOTP for different similarity measures – the exponential cost in black triangle, IoU cost in red square, linear cost in blue asterisk – and different detection types – human detection in solid line, action recognition in dash-dot line, ground truth in dashed line – shown as functions over sample interval.

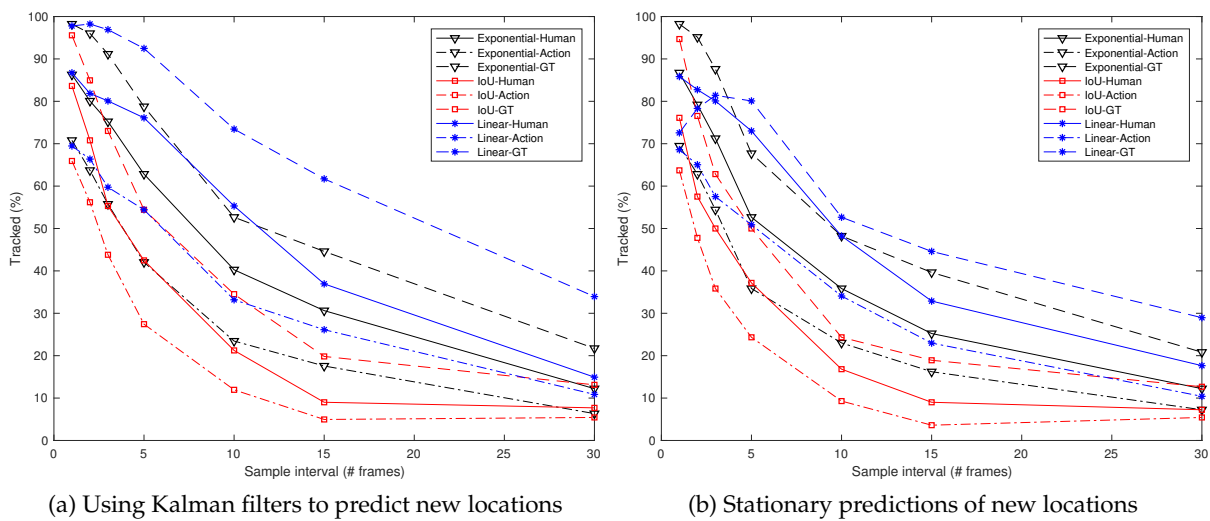

(a) Using Kalman filters to predict new locations     (b) Stationary predictions of new locations

Figure 4.3: Percentage of paths that were *not* considered mostly lost (ML) for different similarity measures – the exponential cost in black triangle, IoU cost in red square, linear cost in blue asterisk – and different detection types – human detection in solid line, action recognition in dash-dot line, ground truth in dashed line – shown as functions over sample interval.

the MOTP started to increase when the frame-rate was low enough, and also that stationary predictions could outperform the use of Kalman filters. When using stationary predictions and ground truth detections as input, all similarity measures had an MOTP close to 100%, for any frame-rate.

### 4.2.1   Real-time tracking

When running the full model on the Jetson TX1 Developer Kit, it could run in 4 FPS when using the human detection model, and in 2 FPS when using the model for action recognition. This corresponds to a sampling interval of 7.5 and 15 respectively. As can be seen in Figures 4.1 to 4.3, the performance at such frame-rates is significantly lower than when using all frames. Figure 4.4 highlights this by showing the MOTA of the best configurations at the sampling intervals corresponding to real-time running. Compared to the performance when analysing all frames, running C++SORT in real-time with the human detection model leads to a drop of almost 50%. When using the action recognition model, the performance drops to a third compared to when using all frames.

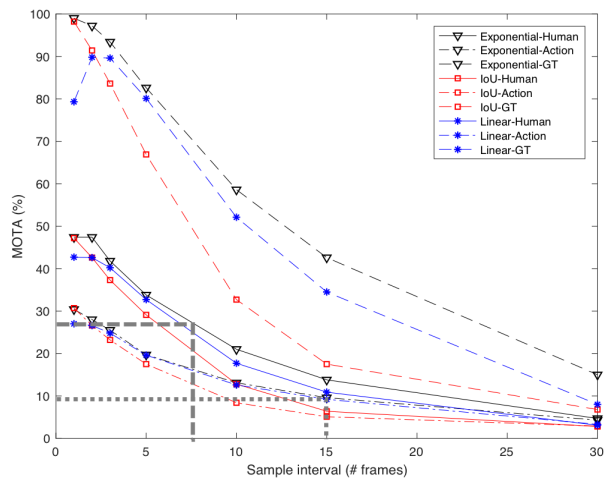

Figure 4.4: MOTA for using the Kalman filter with different similarity measures – the exponential cost in black triangle, IoU cost in red square, linear cost in blue asterisk – and different detection types – human detection in solid line, action recognition in dash-dot line, ground truth in dashed line – shown as functions over sample interval. The dashed and the dotted elbow highlights the performance loss when running C++SORT – with the human detection and the action recognition model respectively – in real-time.

# Chapter 5

# Discussion and conclusions

The thesis is concluded in this chapter. First, a discussion about the method and the results is held in Section 5.1, followed by ethical aspects in Section 5.2 and some concluding remarks in Section 5.3. Last, a few suggestions on extensions and future work is given in Section 5.4.

## 5.1 Method and results

By studying the results in Section 4.1, it is apparent that even a simple approach such as the one described in this report can be competitive, at least when considering the speed. Our method, C++SORT, is by far the fastest algorithm, outpacing the other by factors of $10 \sim 1000$. In fact, there is only one other method, *RNN_LSTM* [35], that runs above 30 FPS, and consequentially no other method could run on Okutama-Action in real-time, even with provided detections. Even *SORT* [25], which is the same method implemented in another language, runs at less than $1/4$ of the speed of C++SORT.

One could argue that the frame-rate C++SORT achieves is unnecessarily high, since video footage is normally captured at around 30 FPS. There are two counter arguments to this statement. First, if one were to run the same models on less powerful hardware, some of the slower algorithms would be impractical to use, while our approach could potentially still run in real-time. This could be the case in *e.g.* small scale robotics, where the use of powerful hardware might not be feasible due to size or budget. Second, and maybe more importantly, the speed displayed on the leaderboards is measuring only the tracker, and not the detector. As discussed in Section 2.3, this might not be indicative of the performance in a real world scenario. If the tracker is to be run in real-time, as video footage is captured, then the detections will not be available as text files. Instead, each frame will need to be processed by an object detector, before being passed to the tracker. On embedded systems, it is likely that the detector and the tracker will share resources, and thus the frame-rate of the tracker will be limited by the speed of the detector. Having a very fast tracking system ensures that the tracker will not add any notable delay to the entire system of detection and tracking. For example, if the object detector runs at 10 FPS, and the tracker runs at 1000 FPS, then the computation time of the tracker is negligible. However, if the tracker also runs at 10 FPS, the effective frame-rate will be 5 FPS.

The results of the experiments on Okutama-Action, with different configurations run at multiple sampling intervals, highlight the importance of a fast tracker. Figures 4.1 to 4.3 in Section 4.2 show a clear drop in performance as more frames are skipped. The case study on the Jetson TX1 Developer Kit shows that running a model in real-time by skipping frames decreases the performance by around 50%, and in some cases up to 90% – *e.g.* the percentage of tracked paths when using the IoU cost with action recognition. Although this experiment was only run on C++SORT, it is believed that a similar drop would occur for other methods as well. One could, however, argue that methods which use appearance features should be more robust to decreased frame-rates, since predicting new locations of objects might be hard at low frame-rates, while appearance features might change less.

From the same experiments, it is also obvious that the detection quality is highly influential on the tracking performance. At full frame-rate, almost every configuration gives near-perfect performance when using the ground truth detections as input. This might not be too surprising, since if the detections are perfect, then no special care has to be taken of false positives or false negatives, and thus a good strategy is to always associate the closest detected objects in two consecutive frames. One can note however, that this is not sufficient to recover the ID of occluded objects once they return, since even perfect detections will not contain information on objects in frames where they are occluded. What is more interesting is that even with perfect detections, the performance is low at low frame-rates. One possible explanation for this is that our method is simply not good enough at predicting new locations of tracked objects. It could also be that low frame-rate tracking is a significantly harder problem, that requires different approaches to solve. For example, the use of appearance features could be the missing component, an option that is not explored in this work.

Some words have to be said about the metric MOTP. As seen in the definition in Equation (2.24), MOTP is determined by how accurate the placement of the bounding boxes of objects are. In tracking-by-detection, this is more related to the detection quality, and not the actual tracking. Thus, having access to the same detections, two different tracking-by-detection models can be expected to get similar MOTP. This is clearly the case in MOTChallenge, where the MOTP of all groups are almost the same, while other metrics vary greatly (see Tables 4.3 and 4.4). This also explains why the plots of MOTP in Figure 4.2 look so different from the other two metrics. Decreasing the frame-rate does not make it more difficult to determine if a detection is correct or not, it only makes it harder to connect detections to long, consistent paths. The reason why stationary predictions in Figure 4.2b outperforms the use of Kalman filters in Figure 4.2a is probably that the stationary predictor trusts the location of the detection fully, whereas the other method uses information from both the Kalman filter and from the detection to determine the true location of the object. But if the detections are accurate – as is the case when using the ground truth – it is better to use only this information instead of adjusting it by a Kalman filter.

The comparison between using Kalman filters and stationary predictions is important to make for the other two metrics as well. In general, the use of Kalman filters leads to increased performance, though not by a large margin, as seen in Figures 4.1 and 4.3. The fact that stationary predictions give similar result even at low frame-rate is surprising; it was believed that as the time in-between frames increased, the need for accurate predictions would make the use of Kalman filters more rewarding. However, as discussed above, it seems that even when using Kalman filters, our model is not able to predict new locations well enough to set it apart from the version using stationary predictions.

As a final remark, it is interesting to note that no similarity measure gets the best performance in more than one metric – the exponential cost results in the highest MOTA, the IoU cost gives the highest MOTP, and the linear cost gets the highest percentage of tracked objects. However, as we made a point earlier that MOTP might not be a very meaningful metric for tracking-by-detection systems, it seems that the IoU cost is the least effective similarity measure, at least at slightly reduced frame-rates. The linear cost, on the other hand, shows a strange behaviour when used with the ground truth, in that it performs poorly a full frame-rate, but well when the sampling interval is increased a few steps. This could be related to the fact that the formula to calculate the linear cost, Equation (3.2), was inverted, which made finding a good threshold value more difficult.

## 5.2  Ethics

The first application of multiple object tracking to come to the reader's mind is probably surveillance. A government or company that is constantly tracking the population is a recurring theme in science fiction books and movies. And while there are potential privacy issues connected with tracking people's movement, there are more effective ways of doing that, such as collecting GPS information from cell phones. Also, real-time performance is not necessarily required for surveillance tasks, unless combined with automatic detection of suspicious behaviour (*e.g.* for crime prevention).

Another use case, where running in real-time can be of the essence, is robotics. To enable robots to navigate in an environment with other moving objects, it can be important for the robot to keep track of those objects, to avoid them or to interact with them. Whether or not it is desirable to have robots understanding their surrounding – or indeed to have robots at all – is a big question, with many arguments for either side. Although having access to better tracking algorithms could lead to the development of dangerous robots, we argue that this is already possible without our work, and in fact without any tracking capabilities.

However, the potential risks of this technology should not be ignored. If that is not already the case, it is likely that tracking systems will be incorporated in future weapons, for missile guidance or even target localisation and assignment. It is unfortunate that much of today's technology can be used in the weapon industry, but we firmly believe that our work – and the field of multiple object tracking – is not connected to the development of weapons. Instead, we hope that fast tracking methods can be used to help people in their daily lives and that the benefits outweigh the potential downsides.

## 5.3  Conclusion

The goal of this work was to implement a fast and competitive tracker, and use it to investigate if real-time multiple object tracking is possible. For this, C++SORT – a reimplementation and slight extension of *SORT* [25] – was created, which when compared to other work available on public leaderboards shows decent performance while being many times faster. By doing a small case study where video was sampled at a low frame-rate to simulate a real-world scenario of real-time tracking, it was seen that the performance reported on MOTChallenge is not representative of the current state of real-time tracking. On certain metrics, the performance dropped as much as 90% when running C++SORT in real-time, and though other models might be more robust under low frame-rates, it is likely that their performance also drops substantially. This is especially likely for the slower trackers, which will have to use a very high sampling interval in order to work in real-time.

The hope is that this work will inspire others to develop faster tracking algorithms, which can run in real-time. Alternatively, since today's CNN based object detectors seem to be a bottleneck in terms of speed, it would be interesting to see work on low frame-rate tracking, thus using the power of CNNs without being limited by their speed. One way to motivate people to work on this would be if MOTChallenge had a task specifically focused on real-time tracking, forcing researchers to develop fast detectors and trackers, or a challenge with low frame-rate video, with some restriction on the speed of the trackers.

## 5.4   Future work

Early on, the decision was made to not use any appearance features in the similarity measure, as to not reduce the speed of the tracker. However, after running the experiments and seeing that none of the three measures used could handle low frame-rate video, it would be interesting to see if this could be solved by using appearance features, or if the problem lies elsewhere in our approach.

It was shown clearly that the detection quality greatly impacts the tracking capabilities of our method, which has been suggested by others to be true for tracking-by-detection models in general [25, 26]. This work was mostly focused on the trade-off between speed and performance of the tracker, but an alternative would be to look at the same thing when choosing an object detector. Increased accuracy for object detectors comes at the cost of decreased speed, and so it is not clear what model to use for real-time tracking. It is possible that there is an optimal balance between performance and speed, that gives good enough detections for the tracker to work with, but does not lead to a too low frame-rate.

# Bibliography

[1] Alex Krizhevsky, Ilya Sutskever, and Geoffrey E Hinton. Imagenet classification with deep convolutional neural networks. In *Advances in neural information processing systems*, pages 1097–1105, 2012.

[2] Olga Russakovsky, Jia Deng, Hao Su, Jonathan Krause, Sanjeev Satheesh, Sean Ma, Zhiheng Huang, Andrej Karpathy, Aditya Khosla, Michael Bernstein, Alexander C. Berg, and Li Fei-Fei. ImageNet Large Scale Visual Recognition Challenge. *International Journal of Computer Vision (IJCV)*, 115(3):211–252, 2015. doi: 10.1007/s11263-015-0816-y.

[3] Kaiming He, Xiangyu Zhang, Shaoqing Ren, and Jian Sun. Delving deep into rectifiers: Surpassing human-level performance on imagenet classification. In *Proceedings of the IEEE international conference on computer vision*, pages 1026–1034, 2015.

[4] Yann LeCun, Yoshua Bengio, and Geoffrey Hinton. Deep learning. *Nature*, 521(7553): 436–444, 2015.

[5] Jürgen Schmidhuber. Deep learning in neural networks: An overview. *Neural networks*, 61:85–117, 2015.

[6] Andrej Karpathy, George Toderici, Sanketh Shetty, Thomas Leung, Rahul Sukthankar, and Li Fei-Fei. Large-scale video classification with convolutional neural networks. In *Proceedings of the IEEE conference on Computer Vision and Pattern Recognition*, pages 1725–1732, 2014.

[7] Jiuxiang Gu, Zhenhua Wang, Jason Kuen, Lianyang Ma, Amir Shahroudy, Bing Shuai, Ting Liu, Xingxing Wang, and Gang Wang. Recent advances in convolutional neural networks. *arXiv preprint arXiv:1512.07108*, 2015.

[8] NASA. Unmanned aircraft system (UAS) traffic management (UTM), 2017. URL `https://utm.arc.nasa.gov/`. Accessed: May 24, 2017.

[9] Jonathan Huang, Vivek Rathod, Chen Sun, Menglong Zhu, Anoop Korattikara, Alireza Fathi, Ian Fischer, Zbigniew Wojna, Yang Song, Sergio Guadarrama, et al. Speed/accuracy trade-offs for modern convolutional object detectors. *arXiv preprint arXiv:1611.10012*, 2016.

[10] MOTChallenge leaderboard, 2016. URL `https://motchallenge.net/results/MOT16/`. Accessed: May 24, 2017.

[11] Mohammadamin Barekatain, Miquel Martí, Hsueh-Fu Shih, Samuel Murray, Kotaro Nakayama, Yutaka Matsuo, and Helmut Prendinger. Okutama-Action: An aerial view video dataset for concurrent human action detection. In *1st Joint BMTT-PETS Workshop on Tracking and Surveillance, CVPR*, pages 1–8. IEEE, 2017.

[12] Arnold WM Smeulders, Dung M Chu, Rita Cucchiara, Simone Calderara, Afshin Dehghan, and Mubarak Shah. Visual tracking: An experimental survey. *IEEE Transactions on Pattern Analysis and Machine Intelligence*, 36(7):1442–1468, 2014.

[13] Hao Jiang, Sidney Fels, and James J Little. A linear programming approach for multiple object tracking. In *Computer Vision and Pattern Recognition, 2007. CVPR'07. IEEE Conference on*, pages 1–8. IEEE, 2007.

[14] Jerome Berclaz, Francois Fleuret, Engin Turetken, and Pascal Fua. Multiple object tracking using k-shortest paths optimization. *IEEE transactions on pattern analysis and machine intelligence*, 33(9):1806–1819, 2011.

[15] Jerome Berclaz, Francois Fleuret, and Pascal Fua. Multiple object tracking using flow linear programming. In *Performance Evaluation of Tracking and Surveillance (PETS-Winter), 2009 Twelfth IEEE International Workshop on*, pages 1–8. IEEE, 2009.

[16] Byungjae Lee, Enkhbayar Erdenee, Songguo Jin, Mi Young Nam, Young Giu Jung, and Phill Kyu Rhee. Multi-class multi-object tracking using changing point detection. In *European Conference on Computer Vision*, pages 68–83. Springer, 2016.

[17] Alper Yilmaz, Omar Javed, and Mubarak Shah. Object tracking: A survey. *Acm computing surveys (CSUR)*, 38(4):13, 2006.

[18] Massimo Piccardi. Background subtraction techniques: a review. In *Systems, man and cybernetics, 2004 IEEE international conference on*, volume 4, pages 3099–3104. IEEE, 2004.

[19] M. Everingham, L. Van Gool, C. K. I. Williams, J. Winn, and A. Zisserman. The PASCAL Visual Object Classes Challenge 2012 (VOC2012) Results. `http://www.pascal-network.org/challenges/VOC/voc2012/workshop/index.html`.

[20] Tsung-Yi Lin, Michael Maire, Serge Belongie, James Hays, Pietro Perona, Deva Ramanan, Piotr Dollár, and C Lawrence Zitnick. Microsoft coco: Common objects in context. In *European Conference on Computer Vision*, pages 740–755. Springer, 2014.

[21] ImageNet leaderboard, 2016. URL `http://image-net.org/challenges/LSVRC/2016/results`. Accessed: May 24, 2017.

[22] PASCAL VOC leaderboard. URL `http://host.robots.ox.ac.uk:8080/leaderboard/`. Accessed: May 24, 2017.

[23] COCO leaderboard. URL `http://mscoco.org/dataset/#detections-leaderboard`. Accessed: May 24, 2017.

[24] Shai Avidan. Support vector tracking. *IEEE transactions on pattern analysis and machine intelligence*, 26(8):1064–1072, 2004.

[25] Alex Bewley, Zongyuan Ge, Lionel Ott, Fabio Ramos, and Ben Upcroft. Simple online and realtime tracking. In *Image Processing (ICIP), 2016 IEEE International Conference on*, pages 3464–3468. IEEE, 2016.

[26] Fengwei Yu, Wenbo Li, Quanquan Li, Yu Liu, Xiaohua Shi, and Junjie Yan. Poi: Multiple object tracking with high performance detection and appearance feature. In *European Conference on Computer Vision*, pages 36–42. Springer, 2016.

[27] Ricardo Sanchez-Matilla, Fabio Poiesi, and Andrea Cavallaro. Online multi-target tracking with strong and weak detections. In *European Conference on Computer Vision*, pages 84–99. Springer, 2016.

[28] Chanho Kim, Fuxin Li, Arridhana Ciptadi, and James M Rehg. Multiple hypothesis tracking revisited. In *Proceedings of the IEEE International Conference on Computer Vision*, pages 4696–4704, 2015.

[29] T Klinger, F Rottensteiner, and C Heipke. Probabilistic multi-person tracking using dynamic bayes networks. *ISPRS Annals of Photogrammetry, Remote Sensing & Spatial Information Sciences*, 2015.

[30] Y. LeCun, B. Boser, J. S. Denker, D. Henderson, R. E. Howard, W. Hubbard, and L. D. Jackel. Handwritten digit recognition with a back-propagation network. In David Touretzky, editor, *Advances in Neural Information Processing Systems (NIPS 1989)*, volume 2, Denver, CO, 1990. Morgan Kaufman.

[31] Yi Li, Kaiming He, Jian Sun, et al. R-fcn: Object detection via region-based fully convolutional networks. In *Advances in Neural Information Processing Systems*, pages 379–387, 2016.

[32] Shaoqing Ren, Kaiming He, Ross Girshick, and Jian Sun. Faster r-cnn: Towards real-time object detection with region proposal networks. In *Advances in neural information processing systems*, pages 91–99, 2015.

[33] Wei Liu, Dragomir Anguelov, Dumitru Erhan, Christian Szegedy, Scott Reed, Cheng-Yang Fu, and Alexander C Berg. Ssd: Single shot multibox detector. In *European Conference on Computer Vision*, pages 21–37. Springer, 2016.

[34] Yu Xiang, Alexandre Alahi, and Silvio Savarese. Learning to track: Online multi-object tracking by decision making. In *Proceedings of the IEEE International Conference on Computer Vision*, pages 4705–4713, 2015.

[35] Anton Milan, Seyed Hamid Rezatofighi, Anthony Dick, Ian Reid, and Konrad Schindler. Online multi-target tracking using recurrent neural networks. *arXiv preprint arXiv:1604.03635*, 2016.

[36] Peter Ondruska and Ingmar Posner. Deep tracking: Seeing beyond seeing using recurrent neural networks. *arXiv preprint arXiv:1602.00991*, 2016.

[37] Nicolai Wojke, Alex Bewley, and Dietrich Paulus. Simple online and realtime tracking with a deep association metric. *arXiv preprint arXiv:1703.07402*, 2017.

[38] Kevin Smith, Daniel Gatica-Perez, and J-M Odobez. Using particles to track varying numbers of interacting people. In *Computer Vision and Pattern Recognition, 2005. CVPR 2005. IEEE Computer Society Conference on*, volume 1, pages 962–969. IEEE, 2005.

[39] Rudolph Emil Kalman. A new approach to linear filtering and prediction problems. *Journal of basic Engineering*, 82(1):35–45, 1960.

[40] Greg Welch and Gary Bishop. An introduction to the kalman filter. 1995.

[41] Christophe Andrieu, Nando De Freitas, Arnaud Doucet, and Michael I Jordan. An introduction to mcmc for machine learning. *Machine learning*, 50(1):5–43, 2003.

[42] David Salmond and Neil Gordon. An introduction to particle filters. *State space and unobserved component models theory and applications*, pages 1–19, 2005.

[43] Harold W Kuhn. The hungarian method for the assignment problem. *Naval research logistics quarterly*, 2(1-2):83–97, 1955.

[44] Jack Edmonds and Richard M Karp. Theoretical improvements in algorithmic efficiency for network flow problems. *Journal of the ACM (JACM)*, 19(2):248–264, 1972.

[45] Cheng-Hao Kuo, Chang Huang, and Ramakant Nevatia. Multi-target tracking by on-line learned discriminative appearance models. In *Computer Vision and Pattern Recognition (CVPR), 2010 IEEE Conference on*, pages 685–692. IEEE, 2010.

[46] Sen M Kuo, Bob H Lee, and Wenshun Tian. *Real-time digital signal processing: fundamentals, implementations and applications*. John Wiley & Sons, 2013.

[47] Alfredo Canziani, Adam Paszke, and Eugenio Culurciello. An analysis of deep neural network models for practical applications. *arXiv preprint arXiv:1605.07678*, 2016.

[48] A. Milan, L. Leal-Taixé, I. Reid, S. Roth, and K. Schindler. MOT16: A benchmark for multi-object tracking. *arXiv:1603.00831 [cs]*, March 2016. URL http://arxiv.org/abs/1603.00831. arXiv: 1603.00831.

[49] Keni Bernardin and Rainer Stiefelhagen. Evaluating multiple object tracking performance: the CLEAR MOT metrics. *EURASIP Journal on Image and Video Processing*, 2008(1):1–10, 2008.

[50] Yangqing Jia, Evan Shelhamer, Jeff Donahue, Sergey Karayev, Jonathan Long, Ross Girshick, Sergio Guadarrama, and Trevor Darrell. Caffe: Convolutional architecture for fast feature embedding. *arXiv preprint arXiv:1408.5093*, 2014.

[51] Davis E. King. Dlib-ml: A machine learning toolkit. *Journal of Machine Learning Research*, 10:1755–1758, 2009.

[52] Min Yang and Yunde Jia. Temporal dynamic appearance modeling for online multi-person tracking. *Computer Vision and Image Understanding*, 153:16–28, 2016.

[53] Amir Sadeghian, Alexandre Alahi, and Silvio Savarese. Tracking the untrackable: Learning to track multiple cues with long-term dependencies. *arXiv preprint arXiv:1701.01909*, 2017.

[54] Ju Hong Yoon, Chang-Ryeol Lee, Ming-Hsuan Yang, and Kuk-Jin Yoon. Online multi-object tracking via structural constraint event aggregation. In *Proceedings of the IEEE Conference on Computer Vision and Pattern Recognition*, pages 1392–1400, 2016.

[55] Seung-Hwan Bae and Kuk-Jin Yoon. Confidence-based data association and dis-criminative deep appearance learning for robust online multi-object tracking. *IEEE Transactions on Pattern Analysis and Machine Intelligence*, 2017.

[56] Hilke Kieritz, Stefan Becker, Wolfgang Hübner, and Michael Arens. Online multi-person tracking using integral channel features. In *Advanced Video and Signal Based Surveillance (AVSS), 2016 13th IEEE International Conference on*, pages 122–130. IEEE, 2016.

[57] Francesco Solera, Simone Calderara, and Rita Cucchiara. Learning to divide and conquer for online multi-target tracking. In *Proceedings of the IEEE International Conference on Computer Vision*, pages 4373–4381, 2015.

[58] Jaeyong Ju, Daehun Kim, Bonhwa Ku, David K Han, and Hanseok Ko. Online multi-object tracking with efficient track drift and fragmentation handling. *JOSA A*, 34(2):280–293, 2017.

[59] Young-min Song and Moongu Jeon. Online multiple object tracking with the hi-erarchically adopted gm-phd filter using motion and appearance. In *Consumer Electronics-Asia (ICCE-Asia), IEEE International Conference on*, pages 1–4. IEEE, 2016.

[60] Low Fagot-Bouquet, Romaric Audigier, Yoann Dhome, and Frédéric Lerasle. Online multi-person tracking based on global sparse collaborative representations. In *Image Processing (ICIP), 2015 IEEE International Conference on*, pages 2414–2418. IEEE, 2015.

[61] Jaeyong Ju, Daehun Kim, Bonhwa Ku, David K Han, and Hanseok Ko. Online multi-person tracking with two-stage data association and online appearance model learning. *IET Computer Vision*, 2016.

[62] Ju Hong Yoon, Ming-Hsuan Yang, Jongwoo Lim, and Kuk-Jin Yoon. Bayesian multi-object tracking using motion context from multiple objects. In *Applications of Computer Vision (WACV), 2015 IEEE Winter Conference on*, pages 33–40. IEEE, 2015.

[63] Seung-Hwan Bae and Kuk-Jin Yoon. Robust online multi-object tracking based on tracklet confidence and online discriminative appearance learning. In *Proceedings of the IEEE Conference on Computer Vision and Pattern Recognition*, pages 1218–1225, 2014.

[64] Yutong Ban, Sileye Ba, Xavier Alameda-Pineda, and Radu Horaud. Tracking multi-ple persons based on a variational bayesian model. In *European Conference on Computer Vision*, pages 52–67. Springer, 2016.

# Appendix A

# Result tables

Found in this appendix are the complete results from the experiments of Section 3.3. The tables are presented as they are given in the DevKit of MOTChallenge. Results from different detection types are separated into sections, with each containing one table for every combination of predictor and similarity measure.

## A.1  Human detection

Results from using the human detection are given below.

| Intv | Rcll | Prcn | FAR | GT | MT | PT | ML | FP | FN | IDs | FM | MOTA | MOTP |
|------|------|------|------|-----|----|-----|-----|------|-------|------|------|------|------|
| 1 | 57.7 | 86.6 | 0.39 | 226 | 43 | 146 | 37 | 6558 | 30994 | 1155 | 2492 | 47.2 | 69.0 |
| 2 | 51.9 | 87.4 | 0.33 | 226 | 25 | 135 | 66 | 2730 | 17654 | 676 | 1395 | 42.6 | 69.2 |
| 3 | 45.6 | 88.0 | 0.27 | 226 | 14 | 111 | 101 | 1524 | 13296 | 513 | 1000 | 37.3 | 69.0 |
| 5 | 36.8 | 88.0 | 0.22 | 226 | 7 | 89 | 130 | 740 | 9273 | 392 | 648 | 29.1 | 68.9 |
| 10 | 18.6 | 86.3 | 0.13 | 226 | 1 | 47 | 178 | 217 | 5978 | 211 | 261 | 12.8 | 69.0 |
| 15 | 10.6 | 83.9 | 0.09 | 222 | 0 | 20 | 202 | 100 | 4372 | 106 | 118 | 6.4 | 69.1 |
| 30 | 5.4 | 80.6 | 0.06 | 221 | 0 | 17 | 204 | 32 | 2326 | 33 | 16 | 2.8 | 70.2 |

Table A.1: Results from using the human detection with Kalman filters and the IoU cost as given by the MOTChallenge DevKit.

| Intv | Rcll | Prcn | FAR | GT | MT | PT | ML | FP | FN | IDs | FM | MOTA | MOTP |
|------|------|------|------|-----|----|-----|-----|------|-------|------|------|------|------|
| 1 | 58.3 | 86.1 | 0.41 | 226 | 54 | 141 | 31 | 6882 | 30595 | 1074 | 2508 | 47.4 | 69.0 |
| 2 | 55.5 | 85.9 | 0.40 | 226 | 33 | 148 | 45 | 3340 | 16332 | 609 | 1561 | 44.7 | 68.9 |
| 3 | 52.5 | 85.6 | 0.39 | 226 | 30 | 140 | 56 | 2168 | 11610 | 463 | 1124 | 41.8 | 68.7 |
| 5 | 44.6 | 83.9 | 0.38 | 226 | 13 | 129 | 84 | 1258 | 8129 | 327 | 750 | 33.8 | 68.4 |
| 10 | 31.2 | 81.5 | 0.31 | 226 | 3 | 88 | 135 | 520 | 5055 | 232 | 423 | 21.0 | 67.5 |
| 15 | 23.0 | 80.0 | 0.25 | 222 | 2 | 66 | 154 | 280 | 3770 | 169 | 233 | 13.8 | 67.0 |
| 30 | 10.3 | 76.9 | 0.14 | 221 | 0 | 27 | 194 | 76 | 2206 | 62 | 58 | 4.7 | 67.9 |

Table A.2: Results from using the human detection with Kalman filters and the exponential cost as given by the MOTChallenge DevKit.

| Intv | Rcll | Prcn | FAR | GT | MT | PT | ML | FP | FN | IDs | FM | MOTA | MOTP |
|------|------|------|-----|-----|-----|-----|-----|------|-------|------|------|------|------|
| 1 | 54.0 | 85.6 | 0.40 | 226 | 37 | 159 | 30 | 6643 | 33767 | 1592 | 3154 | 42.7 | 68.9 |
| 2 | 53.9 | 85.4 | 0.40 | 226 | 30 | 155 | 41 | 3372 | 16899 | 786 | 1723 | 42.6 | 68.8 |
| 3 | 52.1 | 84.5 | 0.42 | 226 | 29 | 152 | 45 | 2331 | 11709 | 592 | 1226 | 40.2 | 68.5 |
| 5 | 46.5 | 81.4 | 0.47 | 226 | 11 | 161 | 54 | 1563 | 7853 | 455 | 861 | 32.7 | 68.2 |
| 10 | 33.0 | 74.0 | 0.51 | 226 | 4 | 121 | 101 | 853 | 4920 | 276 | 470 | 17.7 | 67.3 |
| 15 | 25.2 | 70.8 | 0.46 | 222 | 1 | 81 | 140 | 508 | 3662 | 188 | 281 | 10.9 | 66.6 |
| 30 | 12.6 | 66.8 | 0.28 | 221 | 0 | 33 | 188 | 154 | 2149 | 79 | 84 | 3.1 | 67.5 |

Table A.3: Results from using the human detection with Kalman filters and the linear cost as given by the MOTChallenge DevKit.

| Intv | Rcll | Prcn | FAR | GT | MT | PT | ML | FP | FN | IDs | FM | MOTA | MOTP |
|------|------|------|-----|-----|-----|-----|-----|------|-------|------|------|------|------|
| 1 | 54.4 | 86.3 | 0.38 | 226 | 25 | 147 | 54 | 6330 | 33439 | 1585 | 2854 | 43.6 | 68.9 |
| 2 | 47.0 | 87.4 | 0.30 | 226 | 11 | 119 | 96 | 2486 | 19454 | 841 | 1500 | 37.9 | 69.0 |
| 3 | 41.9 | 87.7 | 0.26 | 226 | 7 | 106 | 113 | 1430 | 14217 | 644 | 1070 | 33.4 | 69.1 |
| 5 | 32.4 | 87.9 | 0.20 | 226 | 5 | 79 | 142 | 658 | 9916 | 433 | 631 | 25.0 | 69.4 |
| 10 | 16.6 | 87.3 | 0.11 | 226 | 1 | 37 | 188 | 178 | 6127 | 185 | 229 | 11.7 | 69.5 |
| 15 | 9.9 | 83.7 | 0.08 | 222 | 0 | 20 | 202 | 94 | 4411 | 105 | 117 | 5.8 | 69.6 |
| 30 | 5.7 | 83.4 | 0.05 | 221 | 0 | 16 | 205 | 28 | 2318 | 31 | 16 | 3.3 | 69.8 |

Table A.4: Results from using the human detection with stationary predictors and the IoU cost as given by the MOTChallenge DevKit.

| Intv | Rcll | Prcn | FAR | GT | MT | PT | ML | FP | FN | IDs | FM | MOTA | MOTP |
|------|------|------|-----|-----|-----|-----|-----|------|-------|------|------|------|------|
| 1 | 57.7 | 85.7 | 0.42 | 226 | 41 | 155 | 30 | 7039 | 31003 | 1309 | 3010 | 46.3 | 68.8 |
| 2 | 54.2 | 86.3 | 0.38 | 226 | 26 | 153 | 47 | 3161 | 16790 | 884 | 1809 | 43.2 | 68.9 |
| 3 | 50.1 | 86.4 | 0.34 | 226 | 17 | 144 | 65 | 1927 | 12195 | 647 | 1290 | 39.6 | 69.0 |
| 5 | 43.4 | 87.1 | 0.28 | 226 | 10 | 109 | 107 | 946 | 8302 | 394 | 782 | 34.3 | 68.9 |
| 10 | 32.1 | 87.9 | 0.19 | 226 | 4 | 77 | 145 | 326 | 4987 | 235 | 367 | 24.5 | 69.3 |
| 15 | 22.3 | 85.2 | 0.17 | 222 | 2 | 54 | 166 | 189 | 3802 | 172 | 220 | 14.9 | 69.4 |
| 30 | 11.3 | 83.8 | 0.10 | 221 | 0 | 27 | 194 | 54 | 2180 | 64 | 55 | 6.5 | 70.1 |

Table A.5: Results from using the human detection with stationary predictors and the exponential cost as given by the MOTChallenge DevKit.

| Intv | Rcll | Prcn | FAR | GT | MT | PT | ML | FP | FN | IDs | FM | MOTA | MOTP |
|------|------|------|-----|-----|-----|-----|-----|------|-------|------|------|------|------|
| 1 | 51.8 | 85.5 | 0.38 | 226 | 29 | 165 | 32 | 6421 | 35375 | 2394 | 3855 | 39.8 | 68.8 |
| 2 | 52.8 | 86.2 | 0.37 | 226 | 29 | 158 | 39 | 3100 | 17309 | 1230 | 2031 | 41.0 | 68.8 |
| 3 | 51.7 | 86.6 | 0.35 | 226 | 21 | 160 | 45 | 1957 | 11802 | 907 | 1460 | 40.0 | 68.9 |
| 5 | 46.7 | 87.1 | 0.30 | 226 | 12 | 153 | 61 | 1013 | 7816 | 622 | 945 | 35.6 | 68.9 |
| 10 | 34.4 | 87.1 | 0.22 | 226 | 6 | 103 | 117 | 373 | 4821 | 332 | 457 | 24.8 | 68.9 |
| 15 | 26.5 | 85.5 | 0.20 | 222 | 2 | 71 | 149 | 221 | 3595 | 214 | 278 | 17.6 | 69.0 |
| 30 | 16.0 | 84.7 | 0.13 | 221 | 1 | 38 | 182 | 71 | 2065 | 89 | 86 | 9.5 | 69.3 |

Table A.6: Results from using the human detection with stationary predictors and the linear cost as given by the MOTChallenge DevKit.

## A.2   Action recognition

Results from using the action recognition are given below.

| Intv | Rcll | Prcn | FAR | GT | MT | PT | ML | FP | FN | IDs | FM | MOTA | MOTP |
|------|------|------|------|-----|-----|-----|-----|------|-------|------|------|------|------|
| 1 | 37.7 | 88.4 | 0.22 | 226 | 16 | 133 | 77 | 3626 | 45679 | 1540 | 2512 | 30.7 | 69.7 |
| 2 | 32.5 | 89.8 | 0.16 | 226 | 7 | 120 | 99 | 1351 | 24742 | 809 | 1363 | 26.6 | 69.8 |
| 3 | 28.5 | 89.9 | 0.14 | 226 | 6 | 93 | 127 | 786 | 17484 | 521 | 862 | 23.2 | 70.0 |
| 5 | 22.0 | 91.0 | 0.10 | 226 | 3 | 59 | 164 | 321 | 11445 | 337 | 509 | 17.5 | 70.2 |
| 10 | 11.3 | 91.1 | 0.05 | 226 | 1 | 26 | 199 | 81 | 6518 | 129 | 153 | 8.4 | 69.9 |
| 15 | 7.2 | 90.5 | 0.03 | 222 | 0 | 11 | 211 | 37 | 4542 | 64 | 66 | 5.1 | 70.1 |
| 30 | 4.1 | 89.5 | 0.02 | 221 | 0 | 12 | 209 | 12 | 2357 | 19 | 10 | 2.9 | 70.3 |

Table A.7: Results from using the action recognition with Kalman filters and the IoU cost as given by the MOTChallenge DevKit.

| Intv | Rcll | Prcn | FAR | GT | MT | PT | ML | FP | FN | IDs | FM | MOTA | MOTP |
|------|------|------|------|-----|-----|-----|-----|------|-------|------|------|------|------|
| 1 | 37.9 | 87.9 | 0.23 | 226 | 16 | 144 | 66 | 3818 | 45558 | 1647 | 2622 | 30.4 | 69.6 |
| 2 | 34.9 | 88.6 | 0.20 | 226 | 10 | 134 | 82 | 1643 | 23886 | 880 | 1553 | 28.0 | 69.5 |
| 3 | 32.2 | 87.8 | 0.20 | 226 | 10 | 116 | 100 | 1092 | 16588 | 532 | 969 | 25.5 | 69.5 |
| 5 | 26.5 | 85.5 | 0.20 | 226 | 5 | 90 | 131 | 660 | 10786 | 345 | 628 | 19.7 | 69.5 |
| 10 | 18.9 | 85.4 | 0.14 | 226 | 1 | 52 | 173 | 237 | 5959 | 185 | 285 | 13.1 | 68.6 |
| 15 | 14.3 | 85.3 | 0.11 | 222 | 0 | 39 | 183 | 121 | 4192 | 111 | 141 | 9.6 | 68.3 |
| 30 | 7.0 | 85.2 | 0.05 | 221 | 0 | 14 | 207 | 30 | 2286 | 37 | 37 | 4.3 | 68.7 |

Table A.8: Results from using the action recognition with Kalman filters and the exponential cost as given by the MOTChallenge DevKit.

| Intv | Rcll | Prcn | FAR | GT | MT | PT | ML | FP | FN | IDs | FM | MOTA | MOTP |
|------|------|------|------|-----|-----|-----|-----|------|-------|------|------|------|------|
| 1 | 34.7 | 87.1 | 0.22 | 226 | 12 | 145 | 69 | 3776 | 47864 | 1907 | 2961 | 27.0 | 69.5 |
| 2 | 34.1 | 87.6 | 0.21 | 226 | 9 | 141 | 76 | 1762 | 24175 | 965 | 1637 | 26.6 | 69.4 |
| 3 | 32.5 | 86.3 | 0.22 | 226 | 11 | 124 | 91 | 1258 | 16513 | 616 | 1052 | 24.8 | 69.3 |
| 5 | 28.5 | 82.3 | 0.27 | 226 | 5 | 118 | 103 | 897 | 10490 | 412 | 713 | 19.6 | 69.0 |
| 10 | 21.3 | 77.8 | 0.27 | 226 | 2 | 73 | 151 | 447 | 5783 | 192 | 316 | 12.6 | 68.3 |
| 15 | 17.3 | 75.9 | 0.24 | 222 | 0 | 58 | 164 | 268 | 4047 | 130 | 189 | 9.2 | 67.4 |
| 30 | 8.8 | 70.8 | 0.16 | 221 | 0 | 24 | 197 | 89 | 2243 | 46 | 49 | 3.3 | 68.0 |

Table A.9: Results from using the action recognition with Kalman filters and the linear cost as given by the MOTChallenge DevKit.

| Intv | Rcll | Prcn | FAR | GT | MT | PT | ML | FP | FN | IDs | FM | MOTA | MOTP |
|------|------|------|-----|-----|-----|-----|-----|------|-------|------|------|------|------|
| 1 | 36.5 | 88.6 | 0.21 | 226 | 12 | 132 | 82 | 3447 | 46552 | 1831 | 2720 | 29.3 | 69.5 |
| 2 | 29.9 | 90.1 | 0.14 | 226 | 2 | 106 | 118 | 1198 | 25720 | 972 | 1428 | 24.0 | 69.8 |
| 3 | 26.0 | 91.1 | 0.11 | 226 | 3 | 78 | 145 | 621 | 18089 | 631 | 872 | 20.9 | 69.9 |
| 5 | 19.2 | 91.0 | 0.08 | 226 | 1 | 54 | 171 | 279 | 11853 | 339 | 471 | 15.0 | 70.2 |
| 10 | 9.7 | 91.4 | 0.04 | 226 | 1 | 20 | 205 | 67 | 6634 | 118 | 136 | 7.2 | 70.1 |
| 15 | 6.2 | 88.7 | 0.04 | 222 | 0 | 8 | 214 | 39 | 4588 | 61 | 60 | 4.2 | 70.0 |
| 30 | 4.1 | 88.7 | 0.02 | 221 | 0 | 12 | 209 | 13 | 2357 | 18 | 8 | 2.9 | 69.8 |

Table A.10: Results from using the action recognition with stationary predictors and the IoU cost as given by the MOTChallenge DevKit.

| Intv | Rcll | Prcn | FAR | GT | MT | PT | ML | FP | FN | IDs | FM | MOTA | MOTP |
|------|------|------|-----|-----|-----|-----|-----|------|-------|------|------|------|------|
| 1 | 37.9 | 88.0 | 0.22 | 226 | 18 | 139 | 69 | 3776 | 45560 | 1675 | 2789 | 30.5 | 69.4 |
| 2 | 34.3 | 89.3 | 0.18 | 226 | 7 | 135 | 84 | 1512 | 24081 | 979 | 1621 | 27.5 | 69.6 |
| 3 | 31.7 | 89.9 | 0.16 | 226 | 11 | 112 | 103 | 868 | 16710 | 644 | 1048 | 25.5 | 69.7 |
| 5 | 26.3 | 90.2 | 0.12 | 226 | 3 | 78 | 145 | 419 | 10814 | 380 | 618 | 20.9 | 70.0 |
| 10 | 18.8 | 92.2 | 0.07 | 226 | 2 | 50 | 174 | 117 | 5969 | 192 | 251 | 14.6 | 70.2 |
| 15 | 13.6 | 90.6 | 0.06 | 222 | 0 | 36 | 186 | 69 | 4229 | 112 | 136 | 9.9 | 69.8 |
| 30 | 7.5 | 90.7 | 0.03 | 221 | 0 | 16 | 205 | 19 | 2274 | 38 | 32 | 5.2 | 70.7 |

Table A.11: Results from using the action recognition with stationary predictors and the exponential cost as given by the MOTChallenge DevKit.

| Intv | Rcll | Prcn | FAR | GT | MT | PT | ML | FP | FN | IDs | FM | MOTA | MOTP |
|------|------|------|-----|-----|-----|-----|-----|------|-------|------|------|------|------|
| 1 | 33.5 | 87.8 | 0.20 | 226 | 15 | 140 | 71 | 3413 | 48802 | 2304 | 3334 | 25.7 | 69.5 |
| 2 | 33.1 | 88.8 | 0.18 | 226 | 10 | 137 | 79 | 1531 | 24529 | 1159 | 1809 | 25.8 | 69.6 |
| 3 | 32.2 | 89.2 | 0.17 | 226 | 14 | 116 | 96 | 951 | 16572 | 775 | 1203 | 25.2 | 69.6 |
| 5 | 29.2 | 89.7 | 0.15 | 226 | 9 | 106 | 111 | 493 | 10386 | 479 | 761 | 22.6 | 69.8 |
| 10 | 22.1 | 90.7 | 0.10 | 226 | 2 | 75 | 149 | 167 | 5723 | 227 | 319 | 16.7 | 69.8 |
| 15 | 17.8 | 90.3 | 0.08 | 222 | 0 | 51 | 171 | 94 | 4022 | 137 | 190 | 13.1 | 69.6 |
| 30 | 10.2 | 89.0 | 0.06 | 221 | 0 | 23 | 198 | 31 | 2208 | 54 | 53 | 6.8 | 70.8 |

Table A.12: Results from using the action recognition with stationary predictors and the linear cost as given by the MOTChallenge DevKit.

## A.3   Ground truth

Results from using the ground truth are given below.

| Intv | Rcll | Prcn | FAR | GT | MT | PT | ML | FP | FN | IDs | FM | MOTA | MOTP |
|------|------|------|-----|-----|-----|-----|-----|-----|------|-----|-----|------|------|
| 1 | 98.3 | 100.0 | 0.00 | 226 | 207 | 9 | 10 | 1 | 1245 | 46 | 38 | 98.2 | 95.1 |
| 2 | 91.7 | 100.0 | 0.00 | 226 | 154 | 38 | 34 | 12 | 3041 | 89 | 81 | 91.4 | 91.6 |
| 3 | 84.3 | 99.9 | 0.00 | 226 | 122 | 43 | 61 | 11 | 3844 | 151 | 135 | 83.6 | 89.0 |
| 5 | 68.9 | 99.8 | 0.01 | 226 | 72 | 51 | 103 | 17 | 4561 | 277 | 244 | 66.9 | 85.8 |
| 10 | 37.3 | 99.9 | 0.00 | 226 | 10 | 68 | 148 | 2 | 4605 | 334 | 298 | 32.7 | 83.6 |
| 15 | 21.9 | 99.9 | 0.00 | 222 | 1 | 43 | 178 | 1 | 3820 | 215 | 180 | 17.5 | 83.8 |
| 30 | 9.5 | 100.0 | 0.00 | 221 | 0 | 29 | 192 | 0 | 2225 | 67 | 34 | 6.8 | 88.9 |

Table A.13: Results from using the ground truth with Kalman filters and the IoU cost as given by the MOTChallenge DevKit.

| Intv | Rcll | Prcn | FAR | GT | MT | PT | ML | FP | FN | IDs | FM | MOTA | MOTP |
|------|------|------|-----|-----|-----|-----|-----|-----|------|-----|-----|------|------|
| 1 | 99.1 | 100.0 | 0.00 | 226 | 217 | 5 | 4 | 13 | 650 | 34 | 25 | 99.0 | 94.9 |
| 2 | 97.5 | 99.8 | 0.01 | 226 | 202 | 15 | 9 | 80 | 909 | 49 | 56 | 97.2 | 90.6 |
| 3 | 94.2 | 99.5 | 0.02 | 226 | 170 | 36 | 20 | 125 | 1413 | 82 | 84 | 93.4 | 87.1 |
| 5 | 84.7 | 98.5 | 0.06 | 226 | 121 | 57 | 48 | 194 | 2239 | 115 | 141 | 82.6 | 82.2 |
| 10 | 63.9 | 96.8 | 0.09 | 226 | 50 | 69 | 107 | 155 | 2655 | 230 | 257 | 58.6 | 76.4 |
| 15 | 49.6 | 95.6 | 0.10 | 222 | 19 | 80 | 123 | 113 | 2464 | 232 | 239 | 42.6 | 74.8 |
| 30 | 21.5 | 95.1 | 0.05 | 221 | 0 | 48 | 173 | 27 | 1930 | 133 | 100 | 15.0 | 76.0 |

Table A.14: Results from using the ground truth with Kalman filters and the exponential cost as given by the MOTChallenge DevKit.

| Intv | Rcll | Prcn | FAR | GT | MT | PT | ML | FP | FN | IDs | FM | MOTA | MOTP |
|------|------|------|-----|-----|-----|-----|-----|-----|-------|-----|------|------|------|
| 1 | 80.9 | 99.6 | 0.01 | 226 | 155 | 66 | 5 | 222 | 13989 | 937 | 2429 | 79.3 | 93.9 |
| 2 | 91.1 | 99.4 | 0.03 | 226 | 196 | 26 | 4 | 217 | 3247 | 274 | 563 | 89.8 | 90.2 |
| 3 | 91.8 | 98.5 | 0.06 | 226 | 172 | 47 | 7 | 331 | 2009 | 209 | 312 | 89.6 | 86.8 |
| 5 | 86.3 | 95.1 | 0.19 | 226 | 119 | 90 | 17 | 652 | 2015 | 251 | 305 | 80.1 | 81.8 |
| 10 | 65.9 | 86.8 | 0.44 | 226 | 48 | 118 | 60 | 737 | 2505 | 280 | 376 | 52.1 | 76.3 |
| 15 | 52.3 | 80.4 | 0.56 | 222 | 10 | 127 | 85 | 623 | 2335 | 246 | 311 | 34.5 | 74.4 |
| 30 | 25.1 | 70.0 | 0.48 | 221 | 0 | 75 | 146 | 264 | 1842 | 156 | 152 | 8.0 | 74.4 |

Table A.15: Results from using the ground truth with Kalman filters and the linear cost as given by the MOTChallenge DevKit.

| Intv | Rcll | Prcn | FAR | GT | MT | PT | ML | FP | FN | IDs | FM | MOTA | MOTP |
|------|------|------|------|-----|-----|-----|-----|-----|------|------|------|------|------|
| 1 | 96.4 | 100.0 | 0.00 | 226 | 183 | 31 | 12 | 0 | 2616 | 121 | 113 | 96.3 | 100.0 |
| 2 | 86.3 | 100.0 | 0.00 | 226 | 129 | 44 | 53 | 0 | 5036 | 131 | 122 | 85.9 | 100.0 |
| 3 | 78.7 | 100.0 | 0.00 | 226 | 90 | 52 | 84 | 0 | 5213 | 186 | 173 | 77.9 | 99.9 |
| 5 | 62.9 | 100.0 | 0.00 | 226 | 40 | 73 | 113 | 0 | 5444 | 272 | 244 | 61.0 | 99.9 |
| 10 | 30.9 | 100.0 | 0.00 | 226 | 7 | 48 | 171 | 0 | 5080 | 289 | 250 | 26.9 | 99.7 |
| 15 | 20.7 | 100.0 | 0.00 | 222 | 1 | 41 | 180 | 0 | 3879 | 185 | 151 | 16.9 | 99.8 |
| 30 | 9.7 | 100.0 | 0.00 | 221 | 0 | 28 | 193 | 0 | 2220 | 68 | 35 | 7.0 | 100.0 |

Table A.16: Results from using the ground truth with stationary predictors and the IoU cost as given by the MOTChallenge DevKit.

| Intv | Rcll | Prcn | FAR | GT | MT | PT | ML | FP | FN | IDs | FM | MOTA | MOTP |
|------|------|------|------|-----|-----|-----|-----|-----|------|------|------|------|------|
| 1 | 99.1 | 100.0 | 0.00 | 226 | 216 | 6 | 4 | 0 | 640 | 27 | 16 | 99.1 | 99.9 |
| 2 | 96.6 | 100.0 | 0.00 | 226 | 184 | 31 | 11 | 0 | 1246 | 84 | 61 | 96.4 | 99.9 |
| 3 | 91.9 | 100.0 | 0.00 | 226 | 150 | 48 | 28 | 0 | 1989 | 136 | 89 | 91.3 | 99.9 |
| 5 | 82.5 | 100.0 | 0.00 | 226 | 114 | 39 | 73 | 0 | 2565 | 110 | 86 | 81.8 | 99.8 |
| 10 | 63.2 | 100.0 | 0.00 | 226 | 45 | 64 | 117 | 0 | 2703 | 191 | 159 | 60.6 | 99.8 |
| 15 | 47.7 | 100.0 | 0.00 | 222 | 15 | 73 | 134 | 0 | 2561 | 205 | 166 | 43.5 | 99.8 |
| 30 | 22.5 | 100.0 | 0.00 | 221 | 2 | 44 | 175 | 0 | 1906 | 127 | 81 | 17.3 | 100.0 |

Table A.17: Results from using the ground truth with stationary predictors and the exponential cost as given by the MOTChallenge DevKit.

| Intv | Rcll | Prcn | FAR | GT | MT | PT | ML | FP | FN | IDs | FM | MOTA | MOTP |
|------|------|------|------|-----|-----|-----|-----|-----|-------|------|------|------|------|
| 1 | 29.5 | 100.0 | 0.00 | 226 | 1 | 163 | 62 | 0 | 51736 | 8156 | 5960 | 18.3 | 99.5 |
| 2 | 36.0 | 100.0 | 0.00 | 226 | 1 | 176 | 49 | 0 | 23460 | 3758 | 3134 | 25.8 | 99.6 |
| 3 | 39.7 | 100.0 | 0.00 | 226 | 1 | 183 | 42 | 0 | 14749 | 2520 | 2193 | 29.4 | 99.6 |
| 5 | 40.9 | 100.0 | 0.00 | 226 | 3 | 178 | 45 | 0 | 8679 | 1433 | 1290 | 31.1 | 99.7 |
| 10 | 33.6 | 100.0 | 0.00 | 226 | 0 | 119 | 107 | 0 | 4881 | 636 | 596 | 24.9 | 99.7 |
| 15 | 33.7 | 100.0 | 0.00 | 222 | 1 | 98 | 123 | 0 | 3242 | 391 | 336 | 25.8 | 99.8 |
| 30 | 22.3 | 100.0 | 0.00 | 221 | 0 | 64 | 157 | 0 | 1910 | 157 | 116 | 15.9 | 99.7 |

Table A.18: Results from using the ground truth with stationary predictors and the linear cost as given by the MOTChallenge DevKit.